%% file: main.tex
\documentclass[11pt,a4paper]{article}

\usepackage{times,latexsym}
\usepackage[acceptedWithA]{tacl2021v1}
\usepackage[utf8]{inputenc}
\usepackage[T1]{fontenc}   
\usepackage{hyperref}       
\usepackage{url}            
\usepackage{booktabs}      
\usepackage{amsfonts}       
\usepackage{nicefrac}       
\usepackage{microtype}      
\usepackage{amsmath}
\usepackage{graphicx}
\usepackage{xcolor}
\usepackage{adjustbox}
\usepackage{subcaption}
\usepackage{amssymb}
\usepackage{algorithm}
\usepackage[german,english]{babel} 
\usepackage{multirow}
\usepackage{float}
\renewcommand{\arraystretch}{1.5}
\usepackage{color, colortbl}
\definecolor{Orange}{rgb}{1,0.90,0.75}
\definecolor{Green}{rgb}{0.9,1,0.87}
\definecolor{Blue}{rgb}{0.67, 0.9, 0.93}
\usepackage{arydshln}
\usepackage{xspace}
\usepackage[most]{tcolorbox}

\newcommand{\xcomet}{\textsc{xCOMET}\xspace}
\newcommand{\xcometmqm}{\textsc{xCOMET-MQM}\xspace}

\definecolor{major_color}{RGB}{234,153,153}
\definecolor{minor_color}{RGB}{255,153,0}
\definecolor{CustomBlue}{RGB}{57,83,191}

\newtcbox{\clustertab}[1]{on line, box align=base, colback={#1},colframe={#1},size=fbox,arc=2pt,top=-1.5pt, bottom=-1.5pt, left=-1.5pt, right=-1.5pt, boxrule=0pt, enlarge left by=1pt}

\newcommand{\firstcluster}{{\footnotesize\clustertab{CustomBlue!60}{1}}}
\newcommand{\secondcluster}{{\footnotesize\clustertab{CustomBlue!40}{2}}}
\newcommand{\thirdcluster}{{\footnotesize\clustertab{CustomBlue!25}{3}}}
\newcommand{\fourthcluster}{{\footnotesize\clustertab{CustomBlue!15}{4}}}
\newcommand{\fifthcluster}
{{\footnotesize\clustertab{CustomBlue!10}{5}}}
\newcommand{\sixthcluster}
{{\footnotesize\clustertab{CustomBlue!8}{6}}}
\newcommand{\seventhcluster}
{{\footnotesize\clustertab{CustomBlue!6}{7}}}

\title{Fine-Grained Reward Optimization for Machine Translation  \\
using Error Severity Mappings}

\author{
\textbf{Miguel Moura Ramos}\thanks{~~Core contributor and corresponding author: \\\url{miguel.moura.ramos@tecnico.ulisboa.pt}}\,\,$^{1,2}$ \quad
\textbf{Tomás Almeida}$^{1}$ \quad
\textbf{Daniel Vareta}$^{1}$ \quad \\
\textbf{Filipe Azevedo}$^{1,2}$ \quad
\textbf{Sweta Agrawal}$^{2}$  \quad
\textbf{Patrick Fernandes}$^{1,2,3}$  \quad
\textbf{André F. T. Martins}$\thanks{~~Work done while at Unbabel.}\,\,^{1,2,4}$ \quad    
\\
$^1$Instituto Superior Técnico, Universidade de Lisboa (ELLIS Unit Lisbon)
\\
$^2$Instituto de Telecomunicações\quad 
$^3$Carnegie Mellon University\quad
$^4$TransPerfect \\
}

\begin{document}

\maketitle

\begin{abstract}
Reinforcement learning (RL) has been proven to be an effective and robust method for training neural machine translation systems, especially when paired with powerful \textit{reward models} that accurately assess translation quality.
However, most research has focused on RL methods that use sentence-level feedback, leading to inefficient learning signals
due to the \textit{reward sparsity} problem \---\ the model receives a single score for the entire sentence.
To address this, we propose a novel approach that leverages fine-grained, token-level quality assessments along with error severity levels using RL methods. Specifically, we use \xcomet, a state-of-the-art quality estimation system, as our token-level reward model.
We conduct experiments on small and large translation datasets with standard encoder-decoder and large language models-based machine translation systems, comparing the impact of sentence-level versus fine-grained reward signals on translation quality.
Our results show that training with token-level rewards improves translation quality across language pairs over baselines according to both automatic and human evaluation. Furthermore, token-level reward optimization improves training stability, evidenced by a steady increase in mean rewards over training epochs. 
\end{abstract}

\input{introduction}

\input{background_related}

\input{method}

\input{experiment}

\input{results}

\input{conclusion}

\section*{Acknowledgments}

We thank the members of SARDINE lab for their useful and constructive comments. This work was supported by the Portuguese Recovery and Resilience Plan through project C645008882-00000055 (Center for Responsible AI), by the EU’s Horizon Europe Research and Innovation Actions (UTTER, contract 101070631), by the project DECOLLAGE (ERC-2022-CoG 101088763), and by Fundação para a Ciência e Tecnologia through contract UIDB/50008/2020.

\bibliographystyle{acl_natbib}
\bibliography{bibliography, anthology}

\appendix
\section{Details of the Severity Assignment Algorithm}
\label{appendix:severity_assignment}

In this section, we provide a detailed description of the severity assignment algorithm used in our approach, focusing on the case of tokens overlapping annotated error spans. Our implementation resolves multiple overlapping error spans by assigning the token the worst severity among all overlapping spans. 
This choice aligns with MQM annotation practices, where the most severe issue in a region governs its quality classification. 

Formally, if a token overlaps spans labeled minor, major, and critical, the token is assigned critical. We avoid averaging or length-weighted schemes to remain fully tokenizer-agnostic. A token is considered affected by a span if it overlaps with it in any way, not only if it is fully contained. This avoids mismatches in cases where tokens are longer than the annotated spans.

Formally, let
\begin{itemize}
    \item $T = \{t_1, \dots, t_n\}$ be the set of tokens,
    \item $S = \{s_1, \dots, s_m\}$ be the set of annotated error spans,
    \item $\sigma(s) \in \{\textnormal{minor} < \textnormal{major} < \textnormal{critical}\}$ denote the severity of span $s$.
\end{itemize}

The severity assigned to token $t_i$ is defined as:
\[
\sigma(t_i) =
\begin{cases}
\max \{ \sigma(s) \mid s \in S,\, t_i \cap s \neq \varnothing \}, \\
\hspace{2cm} \text{if } \exists s \in S : t_i \cap s \neq \varnothing \\[6pt]
\text{None}, \quad \text{otherwise.}
\end{cases}
\]

\paragraph{Example.}  
Suppose the text is tokenized as a single token ``abc'', and error spans are defined as $[..a]$ (minor), $[b]$ (major), and $[c..]$ (critical).
Since the token ``abc'' overlaps with all three spans, its assigned severity is critical.

\section{Details of the Hybrid sRL–tRL Experiment}
\label{appendix:hybrid_experiment}

\begin{table*}[!htbp]
\centering
\footnotesize
\setlength\tabcolsep{4pt}
{\renewcommand{\arraystretch}{1.2}
\begin{tabular}{lcccccc}
    \toprule
    \multirow{2}{*}{\textsc{Method}} & \multicolumn{6}{c}{Metrics} \\
    \cmidrule(lr){2-7}
        & \textsc{BLEU} & \textsc{ChrF} & \textsc{COMET22} & \textsc{xCOMET} & \textsc{BLEURT} & \textsc{CometKiwi-23} \\
    \midrule
    \textsc{Tower} & 42.77 & 62.60 & 71.80 & 88.50 & 68.59 & 67.20 \\
    \quad + SFT & 45.14 & 63.30 & 72.18 & 88.90 & 69.18 & 67.30 \\
    \quad + sRL w/ \textsc{xCOMET-MQM} & 45.58 & 63.76 & 72.41 & 90.20 & 70.55 & 69.70 \\
    \quad + tRL w/ \textsc{xCOMET-MQM} & \textbf{46.92} & \textbf{65.63} & \textbf{74.66} & \textbf{91.90} & \textbf{71.80} & \textbf{71.20} \\
    \quad + hRL w/ \textsc{xCOMET-MQM} & 45.96 & 65.07 & 73.20 & 90.57 & 71.30 & 69.34 \\
    \bottomrule
\end{tabular}}
\caption{Hybrid RL (hRL) results compared to baselines on WMT24 EN$\rightarrow$DE.}
\label{tab:hybrid_results}
\end{table*}

To examine whether the improvements of tRL are primarily driven by long-sequence behavior,
we evaluate a hybrid reinforcement learning (hRL) approach that applies
sentence-level RL (sRL) to short inputs and token-level RL (tRL) to long ones.
Short sentences are defined as those below the average source length in the training
data, and long sentences as those above it.  
This experiment follows our best-performing configuration: the \textsc{Tower} model with \xcometmqm used as the reward signal.
Table~\ref{tab:hybrid_results} reports the results alongside the relevant baselines
from Table~\ref{tab:results_wmt24}.
The hRL model improves over sRL on most metrics but consistently falls short of
tRL, indicating that tRL provides the strongest training signal across sentence lengths
without the added complexity of a hybrid setup.

\section{Analysis of \textsc{chrF} Drop Cases}
\label{appendix:chrF_analysis}

This section provides a focused quantitative and qualitative analysis of translations where \textsc{chrF} decreases but \xcomet improves under token-level RL (tRL).
The goal is to examine whether the observed \textsc{chrF} drop corresponds to genuine translation degradation or reflects a metric mismatch. 
We leverage available human evaluation data for WMT24 EN$\rightarrow$DE, using Direct Assessment (DA) scores as reliable indicators of translation quality.

This analysis stems from the observation that, in some settings, tRL outputs show a noticeable drop in \textsc{chrF} (up to 7 points on IWSLT2017) while simultaneously improving \xcomet and other quality metrics. This raised concerns that such a drop might reflect lexical imprecision or other undesirable artifacts. To investigate, we conducted a detailed error analysis focusing on cases with the largest discrepancies between \textsc{chrF} and \xcomet. Examining these high-divergence examples provides the clearest insight into whether \textsc{chrF} drops correspond to real quality issues or simply reflect a mismatch between metrics.

\subsection{Quantitative Analysis}

\begin{figure}[!htbp]
\centering
\includegraphics[width=\linewidth]{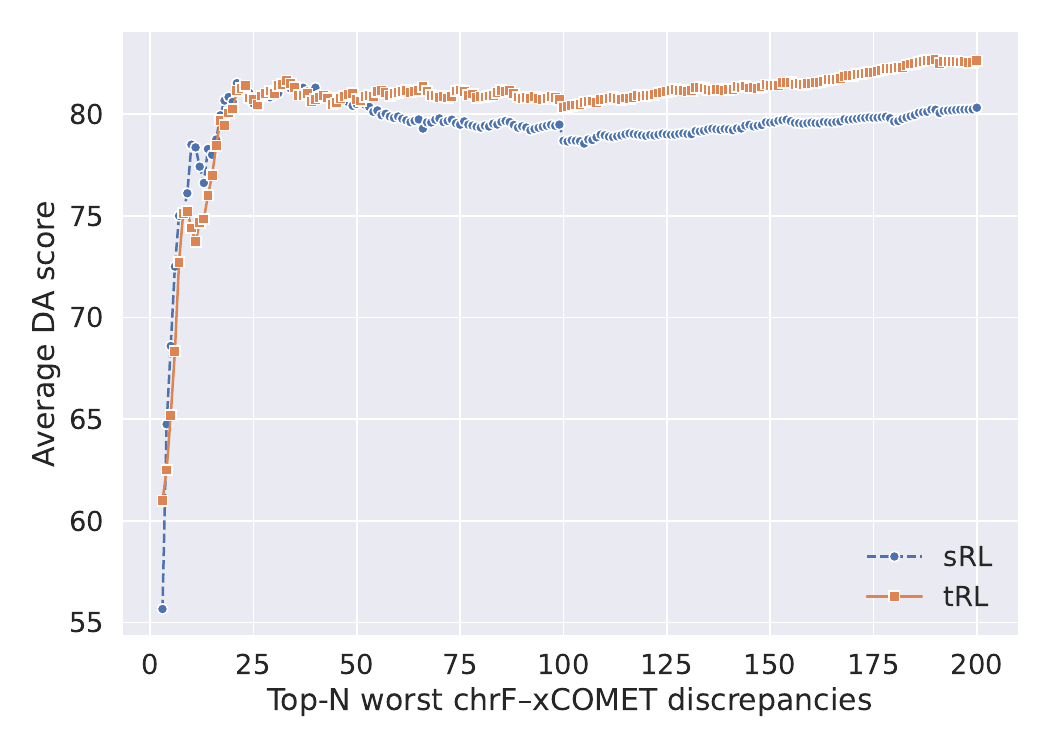}
\caption{Summary of chrF, xCOMET, and DA for cases with extreme metric discrepancies in WMT24 EN$\rightarrow$DE translations.}
\label{fig:chrF_xCOMET_DA}
\end{figure}

Figure~\ref{fig:chrF_xCOMET_DA} summarizes the extreme discrepancy cases for EN$\rightarrow$DE translations. The plot shows the average DA score for the top-N cases with the largest \textsc{chrF}–\xcomet discrepancies. 
Even with the observed \textsc{chrF} drops and corresponding \xcomet increases, tRL consistently achieves higher human DA scores than sRL, indicating that translation quality is not compromised. This strongly suggests that the apparent \textsc{chrF} decline is a metric artifact rather than a genuine degradation in translation quality.

\vspace{1em}

\subsection{Qualitative Analysis}

Table~\ref{tab:appendix_repexamples} presents representative examples of translations with \textsc{chrF} drops but higher \xcomet and DA scores. These cases illustrate how lexical divergence from the reference can lower \textsc{chrF} while yielding translations that are more fluent, semantically accurate, and preferred by human evaluators.

\begin{table*}[!htbp]
    \centering
    \rowcolors{2}{gray!10}{white}
    \scalebox{0.75}{
    \begin{tabular}{l p{\linewidth}}
        \midrule
        \rowcolor{gray!10} \multicolumn{2}{l}{\textbf{Example 1}} \\
        Source & I like both the glasses you posted, these are a really nice colour...makes me want new ones \\
        Reference & Ich mag beide Brillen, die du gepostet hast, die haben eine wirklich nette Farbe ... weckt bei mir den Wunsch, neue zu kaufen \\
        sRL translation & Ich mag beide Brillen, die du gepostet hast, diese sind wirklich eine schöne Farbe... Ich würde gerne neue haben \\
        tRL translation & Mir gefallen beide Brillen, die du gepostet hast, sie haben eine wirklich schöne Farbe... jetzt will ich auch neue \\
        Scores & \textsc{ChrF} 63.4 $\rightarrow$ \textcolor{red}{59.7} \quad \xcomet: 96.8 $\rightarrow$ \textcolor{green!60!black}{99.3} \quad \textsc{DA}: 75.0 $\rightarrow$ \textcolor{green!60!black}{84.0} \\
        \midrule

        \multicolumn{2}{l}{\textbf{Example 2}} \\
        Source & I don't have any telemetry for the battery. This is something I might add in a future revision. \\
        Reference & Ich habe keine Telemetrie für den Akku. So etwas könnte ich in Zukunft noch einbauen. \\
        sRL translation & Ich habe keine Telemetrie für den Akku. Das ist etwas, was ich in einer zukünftigen Überarbeitung hinzufügen könnte. \\
        tRL translation & Ich habe keine Telemetrie-Daten für den Akku. Das ist etwas, was ich vielleicht in einer zukünftigen Version hinzufügen könnte. \\
        Scores & \textsc{ChrF} 58.4 $\rightarrow$ \textcolor{red}{53.1} \quad \xcomet: 97.2 $\rightarrow$ \textcolor{green!60!black}{98.2} \quad \textsc{DA}: 84.0 $\rightarrow$ \textcolor{green!60!black}{94.0} \\
        \midrule

        \multicolumn{2}{l}{\textbf{Example 3}} \\
        Source & “Which notebook is that?” Ivory asked, sitting down next to Kari on her bed. \\
        Reference & \glqq Welches Notizbuch ist das?“, fragte Ivory und setzte sich neben Kari auf deren Bett. \\
        sRL translation & \glqq Welches Notizbuch ist das?“ fragte Ivory und setzte sich neben Kari aufs Bett. \\
        tRL translation & \glqq Welches Notizbuch ist das?“, fragte Ivory, während sie sich neben Kari auf das Bett setzte. \\
        Scores & \textsc{ChrF} 85.3 $\rightarrow$ \textcolor{red}{80.4} \quad \xcomet: 99.3 $\rightarrow$ \textcolor{green!60!black}{99.5} \quad \textsc{DA}: 84.0 $\rightarrow$ \textcolor{green!60!black}{93.0} \\
        \midrule

        \multicolumn{2}{l}{\textbf{Example 4}} \\
        Source & I’m splurging on a new set of frames, these red ones I reeeeally like. \\
        Reference & Ich verschwende mein Geld für eine neue Brillenfassung, diese rote mag ich seeehr. \\
        sRL translation & Ich habe mir eine neue Brille gekauft, diese rote gefällt mir richtig gut. \\
        tRL translation & Ich gönne mir ein neues Brillengestell, diese roten gefallen mir ganz besonders. \\
        Scores & \textsc{ChrF} 37.8 $\rightarrow$ \textcolor{red}{33.8} \quad \xcomet: 95.2 $\rightarrow$ \textcolor{green!60!black}{95.5} \quad \textsc{DA}: 76.0 $\rightarrow$ \textcolor{green!60!black}{92.0} \\
        \midrule

        \multicolumn{2}{l}{\textbf{Example 5}} \\
        Source & “Well my buddys from boot camp here did” Cohren motioned to Harris, Craith and Ravik. “But then again, their squad and platoon leader’s, they know to wake up before their units. \\
        Reference & \glqq Nun, bei meinem Kumpels aus dem Bootcamp ist das bereits der Fall“, Cohren winkte Harris, Craith und Ravik zu. \glqq Aber auch ihre Truppen- und Zugführer wissen, dass sie vor ihren Einheiten aufwachen müssen.“ \\
        sRL translation & \glqq Nun ja, meine Kameraden vom Bootcamp hier haben es getan“, sagte Cohren zu Harris, Craith und Ravik. \glqq Aber ihr Zug- und Platoonführer wissen, dass sie vor ihren Einheiten aufwachen müssen. \\
        tRL translation & \glqq Nun, meine Kameraden vom Bootcamp haben es getan“, sagte Cohren und deutete dabei auf Harris, Craith und Ravik. \glqq Aber sie und ihre Zugführer wissen ja auch, dass sie vor ihren Einheiten aufstehen müssen.“ \\
        Scores & \textsc{ChrF} 57.8 $\rightarrow$ \textcolor{red}{56.2} \quad \xcomet: 81.9 $\rightarrow$ \textcolor{green!60!black}{85.1} \quad \textsc{DA}: 74.0 $\rightarrow$ \textcolor{green!60!black}{85.0} \\
        \bottomrule
    \end{tabular}}
    \caption{Representative translation examples where tRL outputs exhibit lower \textsc{chrF} than sRL but higher \xcomet and human DA scores, illustrating that the \textsc{chrF} drop reflects a metric mismatch rather than an actual decline in translation quality.}
    \label{tab:appendix_repexamples}
\end{table*}

\end{document}

%% file: introduction.tex
\section{Introduction}

\begin{figure*}
    \centering
    \includegraphics[width=\linewidth]{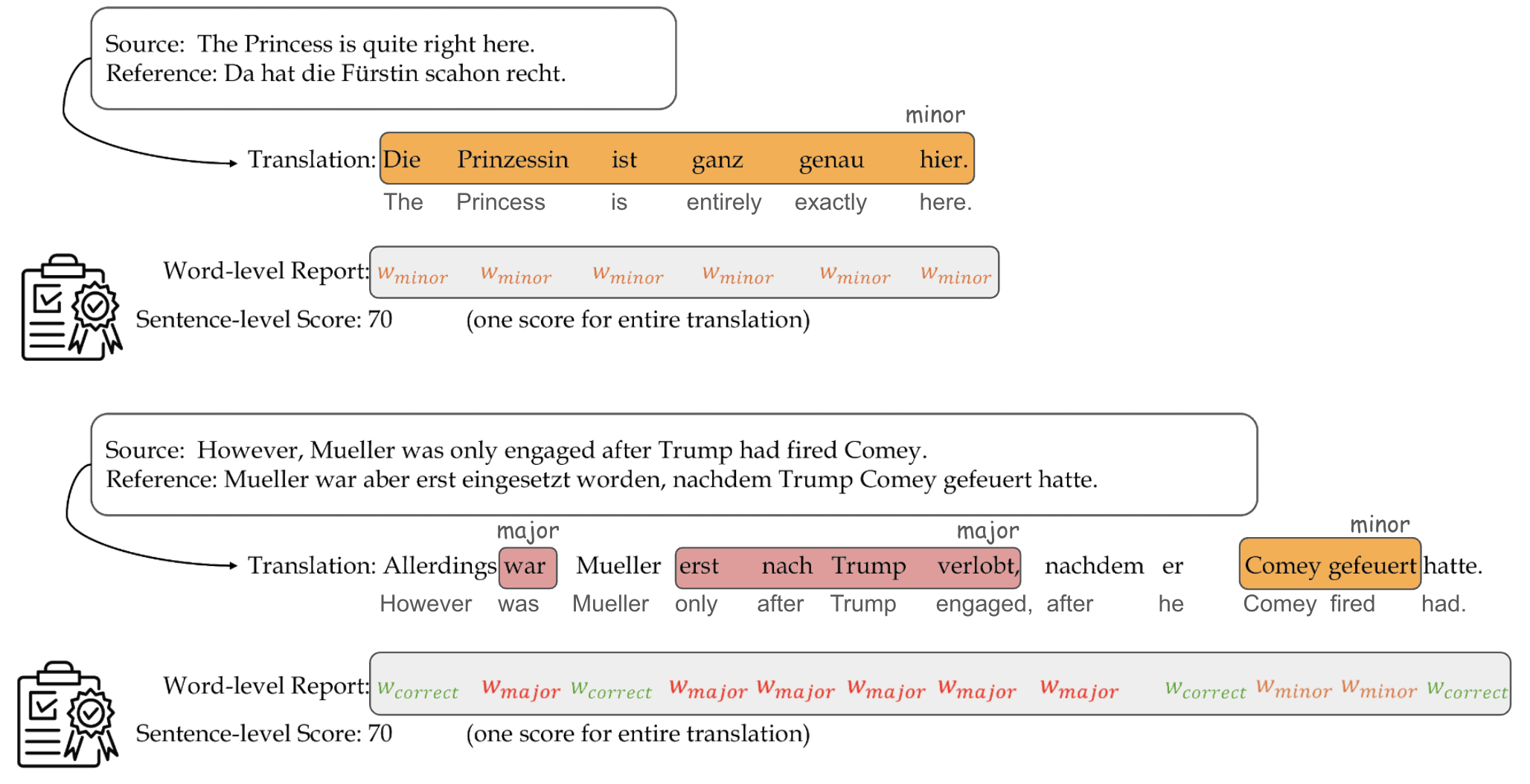}
    \vspace{-1.8em}
    \caption{Two examples are presented, both with identical sentence-level assessments but differing error severity and frequency. The reward model identifies translation error spans along with their corresponding severity levels. In these examples, we highlight both \colorbox{minor_color}{minor} and \colorbox{major_color}{major} error spans. By mapping these spans to numerical values that reflect their severity, we can derive word-level scores/rewards. Since error spans can contain multiple words, we assume that all words within a given span share the same severity. 
    }
    \label{fig:illustration}
\end{figure*}

Neural machine translation (NMT) \cite{Kalchbrenner2013RecurrentCT, sutskever2014sequence, cho2014properties}, a leading approach within MT, leverages neural networks to automate language translation and has driven significant improvements in translation quality. However, most NMT systems are predominantly trained using \textit{maximum likelihood estimation} (MLE). MLE-based training focuses on maximizing the probability of next-word predictions given a partial reference. This often leads to a critical problem known as exposure bias -- the model uses ground-truth prefix tokens during training, but during inference it relies on its previous predictions \cite{BENGIO2015,RANZATO2016,WISEMAN2016}. This can cause errors to propagate through the generated sequence, severely degrading the translation quality. Furthermore, it tends to produce translations that lack global coherence and adequacy as the model does not sufficiently consider the context of entire sentences or the overarching meaning. This has spurred interest in using alternative approaches that leverage RL methods for training NMT systems.

RL-based approaches use explicit reward models to evaluate the outputs generated by the NMT system, assigning scores to generated hypotheses to guide the learning process. However, most prior research \cite{RANZATO2016,wu2016google,Bahdanau2016,nguyen-etal-2017-reinforcement,wuseq2017,kreutzer-etal-2018-neural,kreutzer-etal-2018-reliability,kiegeland-kreutzer-2021-revisiting} predominantly relies on sentence-level feedback and often struggles with \textit{reward sparsity}, particularly for long-form text generation: sentence-level rewards fail to capture specific issues within a translation, making it difficult for the model to learn from negative reward signals. As shown in Figure~\ref{fig:illustration}, two translations corresponding to different source texts of varying length receive the same sentence-level quality score of 70, yet differ significantly in the nature and impact of the errors: The first translation has several minor errors scattered throughout the text, while the latter has major errors that could potentially hinder the understanding of the original content. 
This suggests that learning can be more effective if feedback is provided at a fine-grained level, including precise identification of the nature of errors.  

Recent advancements in automated MT evaluation metrics that generate fine-grained error span predictions, such as \xcomet \cite{xcomet}, \textsc{MetricX} \cite{juraska-etal-2023-metricx}, \textsc{AutoMQM} \cite{fernandes-etal-2023-devil}, \textsc{EAPrompt} \citep{lu-etal-2024-error}, \textsc{MaTESE} \cite{perrella-etal-2022-matese}, and \textsc{BARTScore++} \citep{lu-etal-2023-toward} have shown promise in improving alignment with human translation quality judgments. These metrics directly predict token-level error severity (no error, minor, major, or critical) and optionally provide sentence-level quality assessments or prompt large language models to identify error types (e.g., mistranslation, omission) and severities based on the Multidimensional Quality Metrics (MQM) framework \citep{lommel-mqm}. 

Despite the potential of severity-based metrics to improve translation quality, their application in MT training via RL methods remains relatively underexplored, since it presents several challenges: (i) the feedback, albeit informative and frequent, can be noisy, and (ii) determining the appropriate reward assignments for different severity levels to ensure effective and stable learning is not straightforward.
In this regard, our research aims to answer the following questions:

\begin{enumerate}

\item Does fine-grained RL methods offer benefit over sentence-level feedback in improving translation quality and stabilizing training?

\item Can fine-grained MT metrics be effectively used to provide accurate, detailed, human-aligned feedback to reduce reward sparsity?

\end{enumerate}

When answering these questions, we make the following contributions:

\begin{enumerate}

\item We propose using a fine-grained evaluation metric, \xcomet, to generate token-level rewards, which increases the reward density by providing frequent token-level rewards, thus improving the robustness and stability of RL-based MT. 

\item We introduce a new severity map to effectively use the reward signals, overcoming the limitations of standard MQM scoring, as demonstrated in our experimental results.

\item We conduct experiments on English-to-German (EN$\rightarrow$DE), English-to-French (EN$\rightarrow$FR), German-to-English (DE$\rightarrow$EN), and French-to-English (FR$\rightarrow$EN) translation datasets, comparing the overall translation quality of NMT systems when using sentence and token-level rewards, showing that translation quality improves when employing \xcomet as a reward model.

\end{enumerate}

\noindent
By integrating fine-grained reward signals into NMT training, we demonstrate significant improvements in translation quality and overcome the challenges of exposure bias, reward sparsity, and instability of RL training, paving the way for more reliable and accurate MT systems.

%% file: background_related.tex
\section{Background}

\paragraph{Standard NMT Training.}
NMT systems utilize learnable parameters, denoted as \( \theta \), to estimate the probability distribution \( p_\theta(y|x) \) over a set of possible translations \( \mathcal{Y} \), conditioned on a given source sentence \( x \). 
In the simplest form of NMT training, \textit{maximum likelihood estimation} (MLE) is used, which maximizes the probability of the correct target translation \( y \) given the source sentence \( x \). 
The MLE objective can be expressed as:
\begin{equation}
    \mathcal{L}_{\mathrm{MLE}}(\theta) = \sum_{(x, y) \in D} \log p_\theta(y|x),
\label{eq:MLE}
\end{equation}
where \( D \) represents a dataset of parallel sentences.

\begin{figure*}[!htb]
\centering
\begin{minipage}{\textwidth}
\begin{align}   
    \mathcal{L}_{\mathrm{REINFORCE}}(\theta) &= \mathbb{E}_{\hat{y} \sim p_\theta(y | x)} \left[R(\hat{y}) \log p_\theta(\hat{y} | x)\right] \label{eq:reinforce}
    \\
    \mathcal{L}_{\mathrm{PPO}}(\theta) &= \mathbb{E}_{\hat{y} \sim p_\theta(y | x)} \left[ \min \left\{ \frac{p_\theta(\hat{y} | x)}{p_{\mathrm{old}}(\hat{y} | x)} \hat{A}_{x, \hat{y}}, \, \mathrm{clip}\left(\frac{p_\theta(\hat{y} | x)}{p_{\mathrm{old}}(\hat{y} | x)}, 1-\epsilon, 1+\epsilon\right) \hat{A}_{x, \hat{y}} \right\} \right]
    \label{eq:ppo} 
\end{align}
\end{minipage}
\label{fig:RL_algos}
\vspace*{-2mm}
\caption{Sentence-level RL losses.}
\vspace{-1.0em}
\end{figure*}

\paragraph{Limitations of MLE Training.}
While commonly used in NMT, MLE has several limitations, primarily its weak learning signals from token-level feedback. As MLE assumes gold-reference tokens (teacher-forcing) during training, when exposed to its own incorrect predictions during inference, it can lead to error accumulation and poor performance on longer sequences. Another major limitation is its tendency to optimize for a single ``most likely'' translation, often ignoring the variety of equally valid alternatives, which reduces the model’s ability to generate diverse and natural outputs.
Additionally, MLE is sensitive to noisy or inconsistent reference translations, which can degrade performance by producing unreliable gradient updates. Taken together, these challenges have prompted the exploration of RL methods, which offer more effective feedback on model-generated outputs by optimizing directly for downstream translation quality measures.

\paragraph{Formulating MT as an RL Problem.}
In the context of MT, we can model the translation process as a Markov Decision Process (MDP) \cite{PUTERMAN1990}, defined by the tuple
\((S, A, P, R, \gamma)\) with a finite vocabulary \( \mathcal{V} \).
The state space \( S \) consists of all possible sequences of tokens up to the current time step, which includes the input sequence in the source language, as well as the target language tokens generated so far.
Initially, the state \( s_0 \) corresponds to the input sentence in the source language, \( x = (x_1, x_2, \dots, x_l) \), where each token \( x_i \in \mathcal{V}_{\text{source}} \). At each time step \( t \in [0, T] \), the state \( s_t \) represents the sequence of tokens generated up to that point, which can be expressed as:
\[
s_t = (x_1, x_2, \dots, x_l, \hat{y}_0, \hat{y}_1, \dots, \hat{y}_{t-1})
\]
The agent selects an action \( \hat{y}_t \in A \), which is a token generated by the policy \( p_{\theta} \) based on the current state \( s_t \). The process continues until an end-of-sequence token is generated, completing the translation.
The reference tokens in the target language are denoted by \( y = (y_1, y_2, \dots, y_m) \), where \( y_t \in \mathcal{V}_{\text{target}} \). The generated tokens \( \hat{y}_t \) are evaluated against \( y_t \) to measure the quality of the translation.
For \( t > 0 \), the state transition function \( P : S \times A \to [0, 1] \) defines the probability of transitioning from one state to another by appending a chosen token to the current translation, and the reward function \( R : S \times A \to \mathbb{R} \) assigns a real-valued reward \( r \) to each transition \( (s, \hat{y}) \), where \( s \in S \) and \( \hat{y} \in A \), based on the quality of the generated translation sequence.
Conceptually, the reward function is defined as a mapping from a hypothesis $\hat{y}$ to a score, i.e., $R(\hat{y})$. In practice, many MT metrics additionally condition on the source and/or the reference, which we make explicit as $R(x, \hat{y}, y)$.
Formally, the reward function can be written as:
\[
R(s_{t}, \hat{y}_{t}) = R(x, \hat{y}_{<t}, \hat{y}_{t}, y) = r
\]
Sentence-level rewards are provided only once per translation and evaluate the entire output at once. Token-level rewards, on the other hand, give feedback for each generated token.
The discount factor \( \gamma \in [0, 1] \) is used to weigh future rewards, with \( \gamma = 1 \) typically chosen in MT to ensure that all rewards are valued equally, allowing the optimization of the entire sequence of tokens in the translation rather than focusing on just the initial tokens.
Finally, the goal is to maximize the expected cumulative reward over trajectories \(\hat{y}\) sampled from the \( p_{\theta} \).
The objective function can be written as:
\[
\mathcal{L}_{\text{RL}} = \mathbb{E}_{\hat{y} \sim p_{\theta}} \left[ \sum_{t=0}^{T} R(x, \hat{y}_{<t}, \hat{y}_{t}, y) \right].
\]

\paragraph{Policy Gradient Algorithms}  To optimize the above objective, we can use policy gradient methods:
REINFORCE \cite{WILLIAMS1992,RANZATO2016}, a vanilla policy gradient method, optimizes translation by sampling hypotheses $\hat{y} \sim p_\theta(y | x)$, scoring them with a reward obtained from an MT metric $R(\hat{y})$, and updating the model to maximize expected rewards, as shown in Equation~\ref{eq:reinforce}. 
Despite its simplicity, it often struggles with high variance and instability. 
Proximal Policy Optimization (PPO) \cite{PPO} mitigates this by using a clipped surrogate objective (Equation~\ref{eq:ppo}) to keep policy updates stable within a margin $\epsilon$, efficient and employs Generalized Advantage Estimation (GAE) \cite{schulman2018highdimensionalcontinuouscontrolusing} to compute advantages $\hat{A}$ using rewards $R$ and value function $V$. 
While PPO performs well across various tasks, simpler methods like REINFORCE can sometimes rival or surpass it \cite{ahmadian2024basics}. Both are evaluated in our experiments (\S\ref{sec:results}).

\section{Related Work} 
\label{sec: method}

\paragraph{Advancements and Challenges in Sentence-level Feedback.} 
Incorporating human feedback as rewards and optimizing language models with RL methods effectively aligns them with human preferences \cite{ouyang2022training}, often surpassing MLE.
A notable example in translation tasks is Minimum Risk Training (MRT) \cite{shen2016minimumrisktrainingneural}, which minimizes expected risk based on evaluation metrics to directly improve translation quality.
Recent advances in NMT build on this idea by refining training with feedback from metrics or human evaluations, incorporating alignment techniques and RL methods \cite{nguyen-etal-2017-reinforcement,kreutzer-etal-2018-reliability,wu-etal-2018-study,kiegeland-kreutzer-2021-revisiting, almapref, agrawal-etal-2024-modeling, Zhu2024, he-etal-2024-improving, ramos-aligning}.
Despite these advancements, sentence-level feedback methods face persistent challenges such as sparse rewards, instability, and difficulty handling long sequences \cite{wu-etal-2018-study}. These issues hinder performance, generalization, and robust learning, even with multi-objective optimizations \cite{wu2023finegrained, jang2023personalizedsoupspersonalizedlarge}. 
To address the limitations of sentence-level feedback, recent research has explored finer-grained rewards at the token level for tasks such as language model alignment \cite{xia-etal-2024-inverse, yoon-etal-2024-tlcr, 2023arXiv231104072G, cao-etal-2024-enhancing}, controllable text generation \cite{li-etal-2024-reinforcement}, query generation \cite{2024arXiv241100722O}, among others, but remain relatively underexplored in MT.

\paragraph{Token-level Feedback and Reward Modeling for MT.}
Previous approaches to token-level reward modeling often relied on binary error markings generated by humans \cite{kreutzer-etal-2020-correct, domingo2017segment} or simulated it by comparing model predictions with reference translations based on heuristic methods \cite{petrushkov-etal-2018-learning}. While effective, these methods provide limited feedback due to their binary nature and require costly human annotation, making them less practical for scalable solutions.
Other approaches have employed reward shaping techniques \cite{ng1999policy, wu-etal-2018-study, RS_GOYAL, RS_RATI}, incorporating intermediate rewards along with BLEU \cite{bleu} as the reward function. However, partial BLEU or token-level BLEU are less effective for fine-grained reward modeling, as they depend on exact $N$-gram matching and fail to capture meaningful semantic differences and context.  
Consequently, these methods, while valuable, are limited in their granularity and fail to address the severity of errors introduced at the token level.

%% file: method.tex
\section{Approach}
\label{sec:fgro}

In this section, we present our method for incorporating token-level rewards into RL training for machine translation (MT). To address the limitations of prior approaches, such as binary feedback or coarse sentence-level scores, we use token-level rewards derived from state-of-the-art evaluation metrics that predict error spans and severity levels. These fine-grained signals are then used to guide learning through adaptations of REINFORCE and PPO objectives at the token level, enabling more effective and stable training of MT systems.

\paragraph{Token-level Reward Modeling.}
Building on the MDP formulation for MT, we focus on token-level reward modeling \---\ feedback is provided for individual tokens rather than entire sequences \---\, allowing the model to refine its policy by identifying and addressing specific translation errors. Given an evaluation metric, $\mathcal{M}$ that predicts error spans along with their severity levels (e.g. minor, major, critical) for a hypothesis given source and optionally a gold reference, we assign numerical weights to each token within an error span according to a severity mapping as defined below:
\[
    \text{SEVERITY MAP} = \left\{
        \begin{array}{ll}
            \text{correct span} & : w_\text{correct}, \\
            \text{minor error} & : w_\text{minor}, \\
            \text{major error} & : w_\text{major}, \\
            \text{critical error} & : w_\text{critical}.
        \end{array}
   \right.
\]
We use the evaluation metric, \xcomet as $\mathcal{M}$ as it was shown to achieve the best correlation with human judgments and was the winning submission for the WMT23 Metrics Shared Task \cite{freitag-etal-2023-results}.
The severity weights from \xcomet adhere to the MQM framework \cite{lommel-mqm}, which classifies translation issues into categories such as fluency, adequacy, grammar, and style. 
Each token within an error span is assigned the same severity weight (see Figure~\ref{fig:illustration}). 
We note that although the weights follow the MQM guidelines, they need to be further adjusted depending on the tasks to optimize the performance of token-level RL.

\paragraph{Tokenization-Agnostic Reward Assignment.}
MT systems and evaluation models typically use subword-level tokenization methods such as Byte-Pair Encoding \cite[BPE]{BPE} or SentencePiece \cite{kudo-richardson-2018-sentencepiece}, where words can be split into multiple subword tokens and token boundaries may not align with natural word boundaries.
Given a detokenized hypothesis from the MT system, our evaluation model $\mathcal{M}$ produces error spans defined at the character level. To assign rewards at the token level for the tokenized hypothesis $\hat{y}$, we first re-tokenize the detokenized hypothesis using the same tokenizer applied during model training. This allows us to obtain precise character offsets for each subword token.
We then align tokens to the character-level error spans by checking for overlap: any token whose character span overlaps with an error span inherits the corresponding error severity. This alignment avoids relying on explicit word boundaries or whitespace segmentation, making the reward assignment robust to different tokenization schemes \---\ including those that generate cross-word subword units \---\ and applicable across languages with or without explicit word boundaries.
By grounding token-level rewards in character-level overlap rather than word-based grouping, our method ensures consistency and generalizability across tokenization models and languages.
Finally, if a token overlaps multiple spans, it is assigned the worst severity \---\ critical $>$ major $>$ minor $>$ correct \---\ avoiding averaging or length-weighting to remain tokenizer-agnostic. Formally, for a token $t$ with overlaps $E(t)$, $\ell(t) = \max_{\succ} \{\ell(e) \mid e \in E(t)\};$ 
if $E(t)$ is empty, $\ell(t) = \text{correct}$. This severity is then mapped to its numeric reward (Table~\ref{tab:table_severity_map}). The full algorithm is provided in Appendix~\ref{appendix:severity_assignment}.

\paragraph{Token-level Policy Refinement.}
In token-level RL, we maintain the structure of traditional sentence-level RL losses but adjust them to operate at the token level. We generate the full sequence, compute rewards for each token, and then perform updates for each token separately.
The traditional sentence-level REINFORCE objective (Equation~\ref{eq:reinforce}) is adapted to token-level by adjusting the loss to calculate the reward for each individual token. 
After generating the full sequence, we perform updates for each token one at a time, as follows:

\scriptsize
\begin{equation}
    \mathcal{L}_{RL}(\theta) =
    \mathbb{E}_{\hat{y} \sim p_\theta(y | x)} \left[ \sum_{t=0}^{T}R(x, \hat{y}_{<t+1},y) \log p_\theta(\hat{y}_t | \hat{y}_{<t}, x) \right].
\end{equation}
\normalsize

Here, $R(x, \hat{y}_{<t+1}, y)$ is the reward for token \( \hat{y}_t \), reflecting its contribution to the overall sequence. Similarly, we extend the sentence-level PPO objective (Equation~\ref{eq:ppo}) to token-level by modifying the loss function to compute the policy ratio and advantage for each token independently. The token-level PPO objective is defined as:

\scriptsize
\begin{equation}
    \begin{split}
        \mathcal{L}_{RL}(\theta) = \mathbb{E}_{\hat{y} \sim p_\theta(y | x)} \left[ \sum_{t=0}^{T} \min \left\{ 
        \frac{p_\theta(\hat{y}_t | \hat{y}_{<t}, x)}{p_{\mathrm{old}}(\hat{y}_t | \hat{y}_{<t}, x)} \hat{A}_{x, \hat{y}_{<t}}, \, 
        \right. \right. \\
        \left. \left. \mathrm{clip}\left(\frac{p_\theta(\hat{y}_t | \hat{y}_{<t}, x)}{p_{\mathrm{old}}(\hat{y}_t | \hat{y}_{<t}, x)}, 1-\epsilon, 1+\epsilon\right) \hat{A}_{x, \hat{y}_{<t}} \right\} \right]
    \end{split}
    \label{eq:wppo}
\end{equation}
\normalsize

The policy ratio captures the change in policy for each token \( \hat{y}_t \) relative to the previous policy. 
The token-level advantage is estimated using Generalized Advantage Estimation (GAE) \citep{schulman2018highdimensionalcontinuouscontrolusing}, which balances bias and variance by mixing temporal-difference (TD) errors across multiple steps. The advantage at time step \( t \) is computed as:
\[
A_t = \sum_{l=0}^{T - t - 1} \lambda^l \delta_{t+l},
\]
where \( \delta_t = R(x, \hat{y}_{<t+1},y) + V(x, \hat{y}_{<t+1}) - V(x, \hat{y}_{<t}) \) is the temporal-difference (TD) error, \( \lambda \in [0, 1] \) is the GAE parameter, \( r_t \) is the reward at time step \( t \), and \( V(x, \hat{y}_{<t}) \) is a learned value function that estimates the expected return from the state defined by the input \( x \) and the generated prefix \( \hat{y}_{<t} \). As explained earlier, we consider \( \gamma = 1.0 \) for our use case, and therefore omit the discount factor for notational simplicity.
To ensure stable training, we apply clipping to limit the extent of policy updates, preventing large, unstable shifts. This approach allows for more granular control over the model’s learning, ensuring that each token is generated in a way that maximizes task-specific objectives while maintaining stability in the policy updates.
Clipping also helps mitigate length bias by preventing longer sequences from accumulating disproportionately high rewards.

%% file: experiment.tex
\section{Experiments}

We outline the experiments designed to explore the application of RL for MT, specifically focusing on comparing the impact of sentence-level and token-level reward signals. 

\subsection{Experimental Setup}

\paragraph{Models.}
We use three state-of-the-art models: a standard encoder-decoder MT model, \href{https://huggingface.co/facebook/nllb-200-1.3B}{\textsc{NLLB}} \cite{flores}, and two LLM-based MT systems, \href{https://huggingface.co/Unbabel/TowerInstruct-Mistral-7B-v0.2}{\textsc{Tower}} \cite{tower} and \href{https://huggingface.co/google/gemma-2-9b-it}{\textsc{Gemma}} \cite{gemmateam2024gemma2improvingopen}. 
While \textsc{NLLB} and \textsc{Tower} are dedicated MT models optimized for translation tasks, \textsc{Gemma} is an LLM that exhibits strong multilingual capabilities. 
These models differ in both their architectures and pre-training methodologies. Each is pre-trained on diverse multilingual datasets, establishing them as robust baselines for investigating the effects of SFT and RL techniques.

\paragraph{Data.}
We use the following training datasets in our experiments: (1) The IWSLT2017 dataset \cite{iwslt2017}, with $242$k examples for English-French (EN$\leftrightarrow$FR), supports rapid experimentation and frequent training iterations. (2) The WMT18 dataset \cite{WMT18} contains $42.3$M examples for English-German (EN$\leftrightarrow$DE).  We train \textsc{NLLB} with both datasets and the LLM-based models with (2). Training stops once rewards stabilize, so not all examples are used.
We evaluate \textsc{NLLB} models using their respective test splits: IWSLT17 (EN$\leftrightarrow$FR) and WMT18 (EN$\leftrightarrow$DE). 
To standardize comparison across MT systems (\textsc{NLLB} and LLM-based MT systems), we also evaluate all models on the WMT24 dataset \cite{WMT24}, addressing concerns about data contamination, as \textsc{Tower} training included the WMT18 test set.

\paragraph{Evaluation.}  
We assess translation quality using a comprehensive suite of well-established evaluation metrics. These include \textbf{lexical reference-based metrics}, such as \textsc{BLEU} \cite{bleu} and \textsc{ChrF} \cite{chrf}; \textbf{neural reference-based metrics}, including \textsc{COMET22} \cite{comet22}, \xcomet \cite{xcomet}, and \textsc{BLEURT} \cite{bleurt}; and a \textbf{neural reference-free metric}, \textsc{CometKiwi-23} \cite{cometkiwi}. Lexical metrics focus on word overlap and $N$-gram matching, while neural metrics evaluate translations in terms of semantic coherence and contextual quality. Including a reference-free metric enables evaluation without reliance on predefined reference texts.  
This diverse set of metrics captures multiple dimensions of translation quality, including fluency, grammatical accuracy, semantic adequacy, and contextual relevance. By using various evaluation criteria, we reduce potential biases that may arise from aligning the reward model with a single evaluation metric, ensuring more robust and reliable conclusions about the impact of different approaches on translation quality.  

We apply significance testing at a confidence threshold of $95\%$.
For segment-level metrics, such as \textsc{COMET-22}, we test at the segment level, but for corpus-level metrics, such as \textsc{BLEU} and \textsc{ChrF}, we apply bootstrapping with $100$ samples of size $500$ \cite{koehn-2004-statistical}.
Performance clusters are formed based on statistically significant gaps, and final rankings are derived by averaging the cluster scores across all languages \cite{colombo2022bestsystemsnewperspectives,freitag-etal-2023-results}.
In addition to automated metrics, we conduct human evaluations with two professional annotators, reporting inter-annotator agreement (Pearson's $r$ and Spearman's $\rho$) and $95\%$ confidence intervals across length buckets to assess the reliability of model differences.

\paragraph{Reward Models.} 
We utilize two reward models based on \href{https://huggingface.co/Unbabel/XCOMET-XL}{\xcomet}. 
The MQM-derived reward signal, referred to as \textsc{xCOMET-MQM}, is generated from error span predictions that identify and classify translation errors by severity. 
Token-level reward signals are directly computed from these error spans, while sentence-level rewards are obtained as a weighted average of the token-level severity spans. 
We also use the standard sentence-level reward signal provided by \textsc{xCOMET} as a baseline for comparison.

\paragraph{Training Configurations.}
We finetune \textsc{NLLB}, \textsc{Tower} and \textsc{Gemma} models using MLE and RL methods detailed in Section \ref{sec: method} with the following configurations:
\begin{itemize}
    \item \textbf{SFT:} a baseline model supervised finetuned on the parallel data using MLE \eqref{eq:MLE}. 
    
    \item \textbf{sRL:} We compare using sentence-level xCOMET with BLEU. The learning algorithm used is PPO \eqref{eq:ppo}, a current state-of-the-art alignment method for MT.
    
    \item \textbf{tRL:} We use token-level xCOMET and compare it with partial BLEU, which is based on reward shaping as detailed in Section \ref{sec:fgro}. 
    The learning algorithm used is tPPO \eqref{eq:wppo}, the proposed token-level version of PPO.
    
    \item \textbf{CPO} \cite{CPO}: a state-of-the-art preference optimization learning method for MT, offering a more efficient variant of DPO \cite{DPO}. We construct the preference dataset by generating multiple outputs from the MT model using the training datasets,\footnote{We generate 16 samples with the value of top\_p  top\_k set to 0.9 and 50 respectively.} and then induce preferences using the \xcomet metric, comparing these outputs to human-written references.
\end{itemize}

\paragraph{Hyperparameter Details.} 
We use the same hyperparameter settings for \textsc{NLLB}, \textsc{Tower}, and \textsc{Gemma}.
We use HuggingFace's Transformers library \cite{wolf-etal-2020-transformers} and the \href{https://github.com/huggingface/trl}{Transformers Reinforcement Learning (TRL)} library  to facilitate RL training.
We perform MLE training with Adam \cite{kingma2017adam} as the optimization algorithm, learning rate decay starting from $1 \times 10^{-5}$ and early stopping.
We use PPO with a learning rate of $1.41 \times 10^{-6}$, $\gamma$ set as $0.99$, trajectory limit set as $10,000$.
Mini-batch updates are performed with a batch size of $16$ over $4$ PPO epochs. The translation prompt for LLM-based MT systems is shown in Table~\ref{tab:prompt}.

\begin{table}[!ht]
\centering
\footnotesize
\begin{tabular}{p{7.2cm}}
\toprule
Translate the following text from {\textbf{\{source\_lang\}}} into {\textbf{\{target\_lang\}}}. \\
{\textbf{\{source\_lang\}}}: {\textbf{\{source\_sentence\}}}. \\
{\textbf{\{target\_lang\}}}: \\
\bottomrule
\end{tabular}
\caption{Prompt used for \textsc{Tower} and \textsc{Gemma}.}
\label{tab:prompt}
\vspace{-1.0em}
\end{table}

%% file: results.tex
\subsection{Results and Main Findings}
\label{sec:results}

\input{tables/main_results}

We present the main results of comparing the different methods across datasets trained using NLLB in Table~\ref{tab:results} and across models in Table~\ref{tab:results_wmt24}.

\paragraph{tRL consistently outperforms SFT and sRL methods across neural metrics.} For all translation directions reported in Table~\ref{tab:results}, ``tRL w/ xCOMET-MQM`` outperforms SFT and its sentence-level counterpart ``sRL w/ xCOMET-MQM`` across all neural metrics considered. SFT significantly improves translation quality by tailoring the pre-trained MT model to the specific target language pairs. Moreover, applying RL methods (sRL or tRL) on top of SFT further enhances the MT model's performance by directly optimizing translations based on targeted reward signals. When comparing the sRL and tRL methods, we observe that sRL leads to moderate improvements over SFT, while the gains obtained by tRL are more substantial, particularly when assessed with advanced neural metrics. 
Although the chrF scores for tRL models trained with \xcometmqm are lower on IWSLT2017, this is likely due to a mismatch between the reward signal and the evaluation metric: token-level rewards optimize semantic quality, not character-level overlap. 
This can lead to fluent, accurate translations that diverge lexically from references, reducing chrF despite improved overall quality. 
Neural metrics, which are more robust to surface-level variation and better aligned with human judgments \cite{freitag-etal-2022-results,freitag-etal-2023-results}, as well as human evaluation (see Section \ref{subsec:he}) \---\ the gold standard for assessing translation quality \---\ consistently show improvements with our tRL approach.
Appendix~\ref{appendix:chrF_analysis} presents a focused quantitative and qualitative analysis of cases in which chrF decreases while xCOMET improves.

\input{tables/llm_results}

\paragraph{tRL improves translation quality for LLM-based MT systems, \textsc{Tower} and \textsc{Gemma}.} 
Our severity-based, fine-grained mechanism significantly improves translation quality across all automatic evaluation metrics, as shown in Table~\ref{tab:results_wmt24}.
These findings highlight that tRL not only improves the quality of state-of-the-art MT models but can also significantly boost stronger LLM-based MT systems, demonstrating its broad applicability and potential for advancing multilingual MT systems.

\paragraph{On-policy PPO results in better translation quality than RL-free method, CPO. }
Both sentence-level and token-level RL methods achieve higher evaluation scores than CPO, demonstrating significant improvement in translation quality across language pairs.
Unlike CPO, which focuses on maintaining predefined constraints imposed by the preference dataset, RL methods like PPO can flexibly and dynamically adjust the MT model based on real-time feedback from the reward models via iterative feedback and refinement. 
Furthermore, tRL uses fine-grained reward signals from xCOMET-MQM capturing a wider range of linguistic features and quality indicators, thus offering more precise and contextually relevant feedback during the training process. This feedback can be leveraged more effectively with PPO than with CPO. 

\paragraph{\xcomet is a superior reward model than  lexical MT metrics.}
The role of the reward model in achieving alignment is crucial, as also evidenced by our findings (Tables~\ref{tab:results} and \ref{tab:results_wmt24}).
Our results clearly show that using \xcomet as a reward model, particularly at the token level, significantly improves translation quality as measured by several metrics. 
Given that \xcomet exhibits a strong correlation with human judgments, it proves to be an essential tool for guiding MT models toward higher translation quality. 
In contrast, traditional metrics like BLEU, based on $N$-gram overlap, can fall short in aligning with human judgments as they do not capture contextual nuances and semantic understanding \cite{freitag-etal-2022-results}. Consequently, BLEU performs less effectively compared to neural metrics like \xcomet in this setup which use contextual embeddings. 
Therefore, incorporating neural metrics as reward models is crucial for capturing the subtleties of language and improving the overall quality and reliability of MT models.

\subsection{Ablation Study} \label{subsec:ablation}
We present several ablations to study how the design choices employed impact the learning and the final translation quality of the optimized model. 

\input{tables/severity_map}
\input{tables/results_severity_map}

\paragraph{Choice of severity map impacts learning.}
We investigate the impact of different severity maps on token-level RL training using xCOMET, as detailed in Table \ref{tab:table_severity_map}. 
The severity maps we evaluate include the default MQM-based map (\textsc{MQM}), our custom map (\textsc{Our}), the reversed MQM-based map (\textsc{rMQM}), the reversed custom map (\textsc{rOur}), and a binary map (\textsc{Bin}).
Our findings, shown in Table~\ref{tab:results_severity_map}, highlight the importance of having gradual transitions between reward values. Smooth transition severity map to result in better translation quality. In contrast, abrupt changes in the reward signal can destabilize learning, leading to inconsistent training, oscillations, or convergence to suboptimal policies \cite{RANZATO2016,sutton2018reinforcement}.
Additionally, we find that the binary severity map, which ignores the severity of errors, provides less informative feedback to the model, resulting in slightly lower performance than maps that offer more nuanced assessments.
Although designing custom severity maps can increase complexity and require hyperparameter tuning, our experiments suggest they can be set in a straightforward way with minimal overhead. We leave a more systematic investigation of these mappings, including the possibility of learning them during training, to future work.

\begin{figure}[!ht]
    \centering
    \begin{subfigure}{\linewidth}
        \centering
        \includegraphics[width=\linewidth]{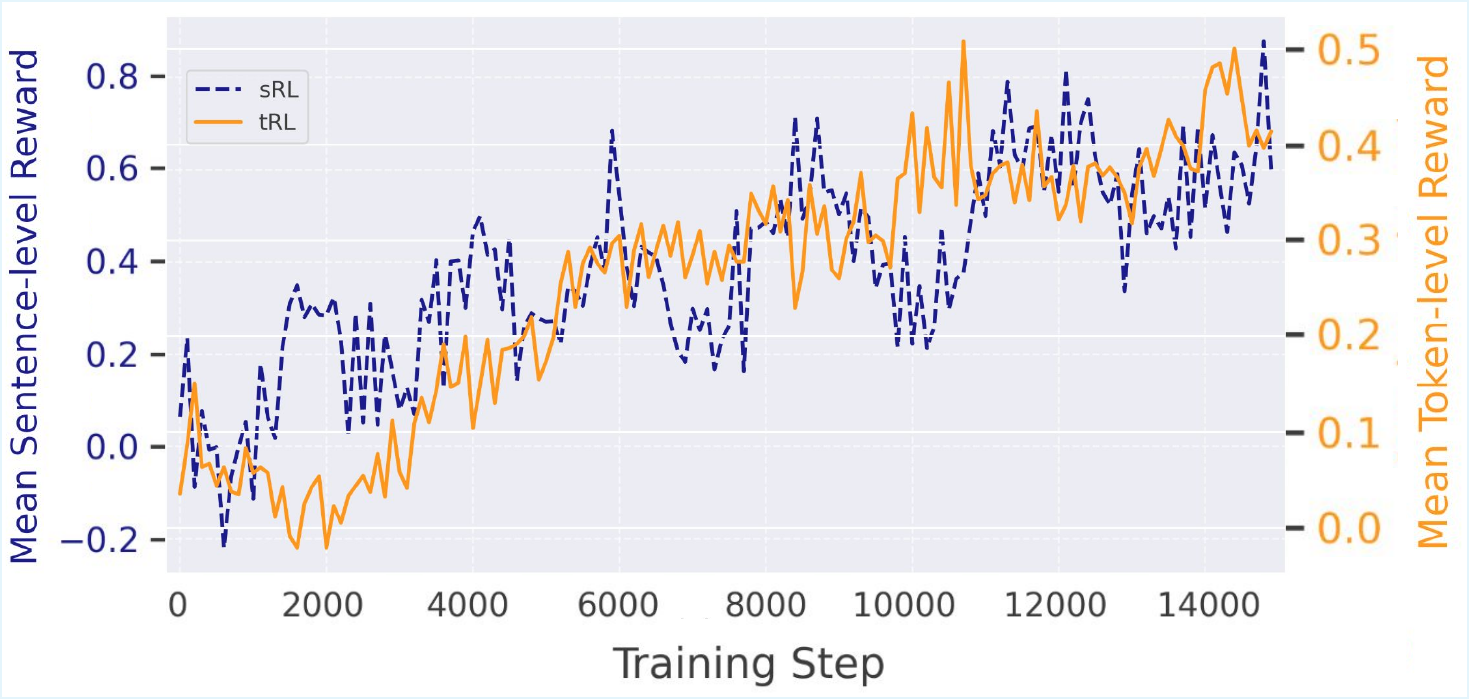}
    \end{subfigure}
    \vspace{0.5em}
    \begin{subfigure}{\linewidth}
        \centering
        \includegraphics[width=\linewidth]{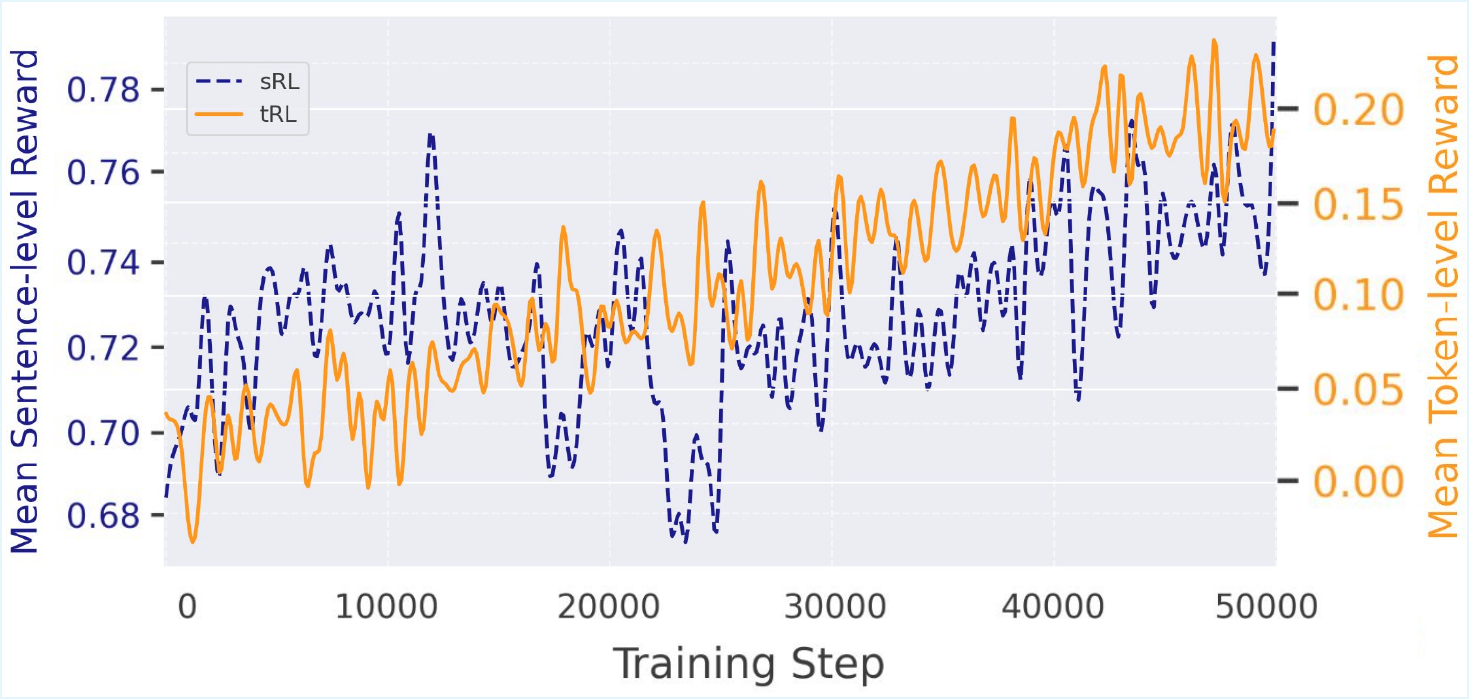}
    \end{subfigure}
    \vspace{-1.5em}
    \caption{Mean rewards per training step for the IWSLT2017 EN→FR (top) and WMT18 EN→DE (bottom) datasets using \xcomet as the reward model with NLLB. The learning curves highlight training stability trends, where tRL (orange) displays greater stability than sRL (blue). Note that reward scales are not directly comparable due to differences in granularity and clipping methods.}
    \vspace{-1em}
    \label{fig:mean_reward}
\end{figure}

\paragraph{tRL improves training stability over sRL.} Figure~\ref{fig:mean_reward} shows the evolution of mean rewards during training for sRL and tRL across the two datasets for the NLLB system. As observed, tRL training exhibits a more stable and consistently increasing reward trajectory, which is crucial for ensuring steady improvements and reducing the risk of performance-degrading fluctuations or overfitting.

\begin{figure}[!ht]
    \centering
    \resizebox{\linewidth}{!}{
        \begin{minipage}{\linewidth}
            \centering
            \includegraphics[width=\linewidth]{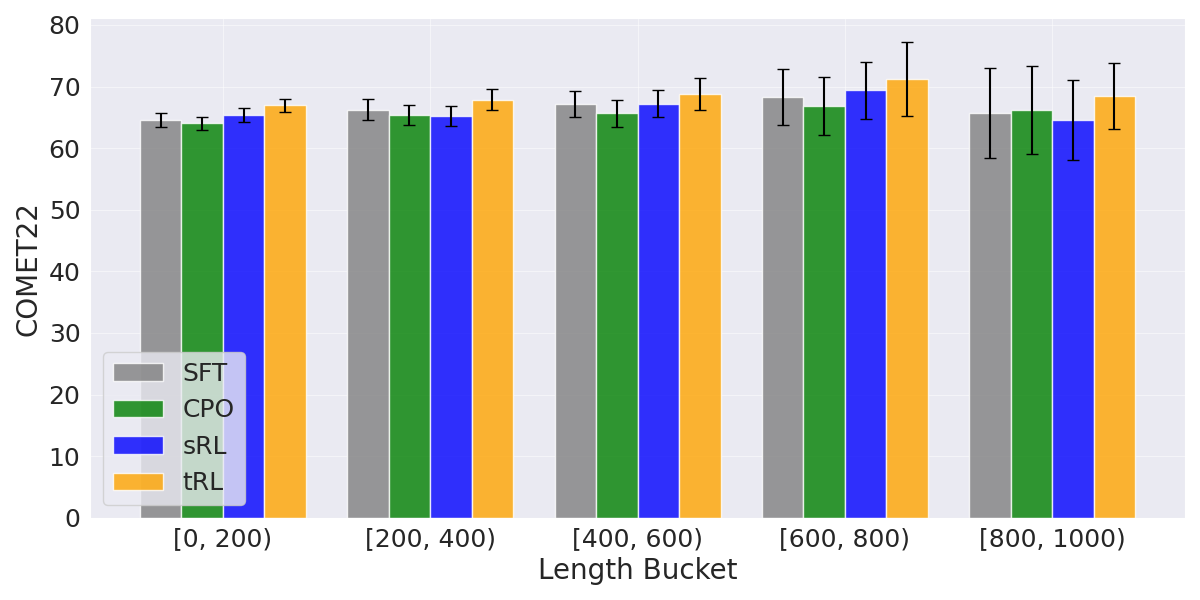}

            \includegraphics[width=\linewidth]{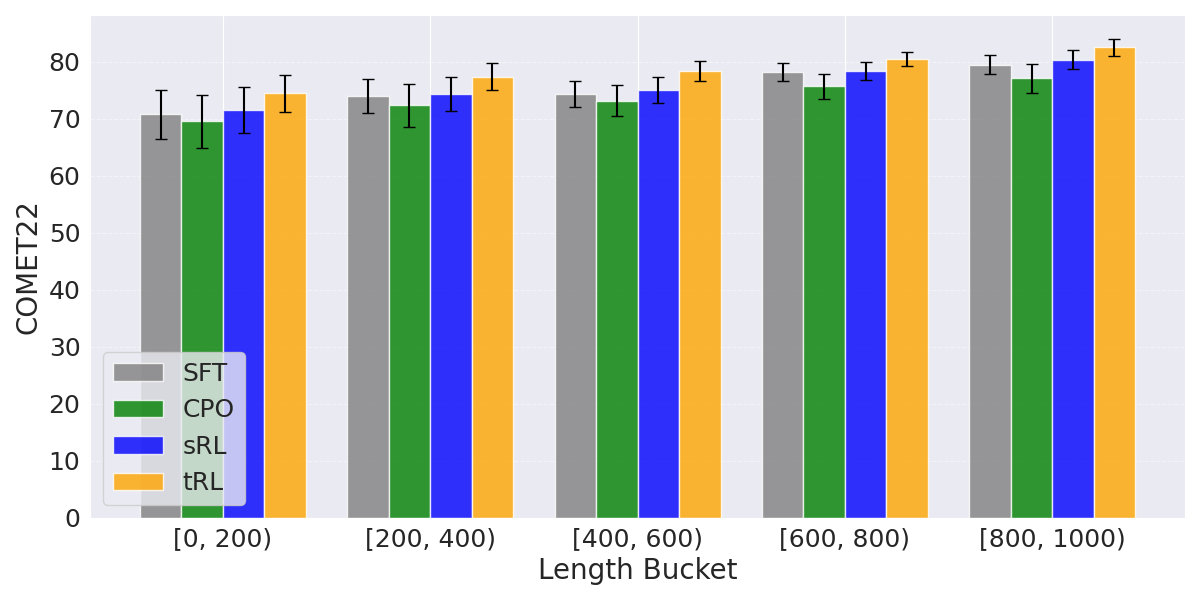}

            \includegraphics[width=\linewidth]{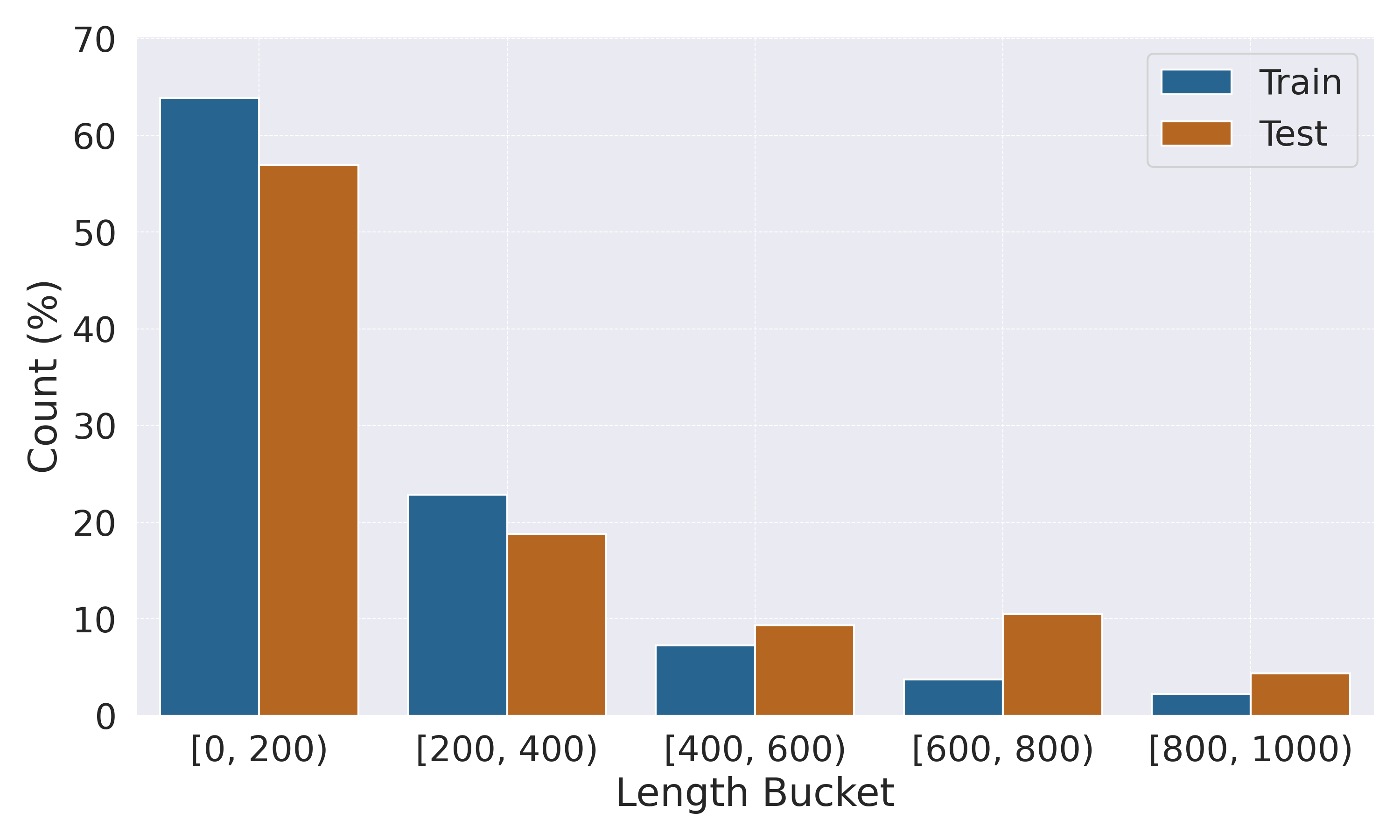}
        \end{minipage}
    }
    \caption{COMET22 scores for \textsc{NLLB} (top), \textsc{Tower} (middle), and a comparative analysis of training and test data length distribution (bottom) on WMT24 EN→DE across increasing source sentence lengths, measured by character string length.}
    \label{fig:fganalysis}
    \vspace{-0.75em}
\end{figure}

\input{tables/results_rl_method}

\paragraph{tRL improves translation quality for longer sequences.} 
Building on our hypothesis that tRL is particularly effective for longer sequences, we present COMET22 scores for the WMT24 EN$\rightarrow$DE dataset in Figure~\ref{fig:fganalysis}, grouped by source sequence length and comparing different training methods, including \textsc{NLLB} and \textsc{Tower}. The figure also shows the distribution of source sequence lengths in the training and test data. Notably, the training data is skewed toward shorter inputs—a common characteristic of large MT corpora—whereas the WMT24 test set includes a broader distribution with a higher proportion of long sequences. This discrepancy highlights the need for models that generalize well to longer inputs.
In this context, "longer sequences" refers to short paragraphs rather than single sentences. These are not document-level inputs and remain within the 512-token limit of \xcomet, ensuring that the reward model processes them without truncation during training.
tRL consistently outperforms other training methods, especially on longer sequences. We attribute this to its ability to capture localized, fine-grained reward signals during training. These findings further support our earlier results: tRL demonstrates the most robust performance, with smaller performance drops as the source sentence length increases, confirming its strength in handling complex sentence structures.
For completeness, we also evaluate in Appendix~\ref{appendix:hybrid_experiment} a hybrid model that applies sRL to short inputs and tRL to long ones. 
This approach outperforms sRL but remains inferior to tRL.

\paragraph{REINFORCE and PPO are suitable methods for training MT systems.} We compare REINFORCE and PPO for RL-based MT with \xcomet as the reward model, evaluating their impact on translation quality (Table \ref{tab:results_rl_method}). Both methods are effective, but PPO achieves superior overall metric scores due to features such as objective clipping and KL divergence control, which enhance training stability. However, REINFORCE remains a strong alternative for simpler implementations that aim to achieve competitive performance.

\paragraph{Efficiency and Quality Tradeoffs in Token-Level Reward Computation.}
Using \xcomet as a reward model for tRL yields higher-quality translations than BLEU, but it also incurs increased computational costs, as detailed in Table \ref{tab:rm_comparison}. Due to its larger pre-trained encoder, \xcomet exhibits significantly higher latency (average seconds per token) and lower throughput (tokens per second). It is worth noting that throughput values can appear higher than latency alone might suggest, as this metric benefits from amortized per-call overheads and batching. The quality improvement, measured by COMET22, reflects the performance of our best model (\textsc{Tower}) on WMT24 EN$\rightarrow$DE after fine-grained optimization with each reward model. Despite the slower processing, \xcomet's superior reward quality is particularly valuable in token-level feedback scenarios where high-quality rewards are essential and the additional computational cost is manageable with GPU acceleration. Furthermore, advances in computational efficiency for transformer architectures—such as quantization \cite{quant}, FlashAttention \cite{fa,fa2}, and distillation \cite{dist}—can help mitigate this computational load, making \xcomet more practical for broader applications.

\begin{table}[!ht]
\centering
\footnotesize
\begin{tabular}{lccc}
\toprule
& \textbf{Latency $\downarrow$} & \textbf{Throughput $\uparrow$} & \textbf{Quality $\uparrow$} \\ \midrule
\textbf{BLEU} & $2.08 \times 10^{-4}$ & $2.24 \times 10^{5}$ & $72.22$ \\
\textbf{\xcomet} & $8.17 \times 10^{-2}$ & $7.64 \times 10^{2}$ & $74.66$ \\
\bottomrule
\end{tabular}
\caption{Comparison of BLEU and \xcomet reward models with respect to computational efficiency (latency, throughput) and final translation quality measured by COMET22.}
\label{tab:rm_comparison}
\end{table}

\subsection{Human Evaluation}
\label{subsec:he}
\paragraph{Setup.} 
For our human evaluation, we used Direct Assessments \cite[DAs]{graham-etal-2013-continuous} to score translations on a scale from 0 to 100, following the standard WMT human evaluation methodology. We evaluate 200 randomly chosen instances from the WMT18 EN $\rightarrow$ DE dataset. Two professional translators, both native speakers of the target language, assess the references and NLLB translations using the following methods: SFT, CPO with \xcomet, tRL with \xcomet, our proposed severity mapping, and sRL with \xcomet.

While WMT18 EN $\rightarrow$ DE allows for easy comparison between several methods due to shorter sequences, we conducted a second human evaluation on the WMT24 EN $\rightarrow$ DE dataset to validate our empirical finding that tRL benefits longer input sequences. For this setting, we directly compare sRL with \xcomet and tRL with \xcomet in a pairwise setting using outputs from the \textsc{Tower} model. To ensure adequate coverage across different sequence lengths, we performed stratified sampling based on source length, using bins [0, 100, 250, 500, 1000] with 50 instances per bin.

\input{tables/human_eval}

\begin{figure}[!ht]
    \centering
    \includegraphics[width=\linewidth]{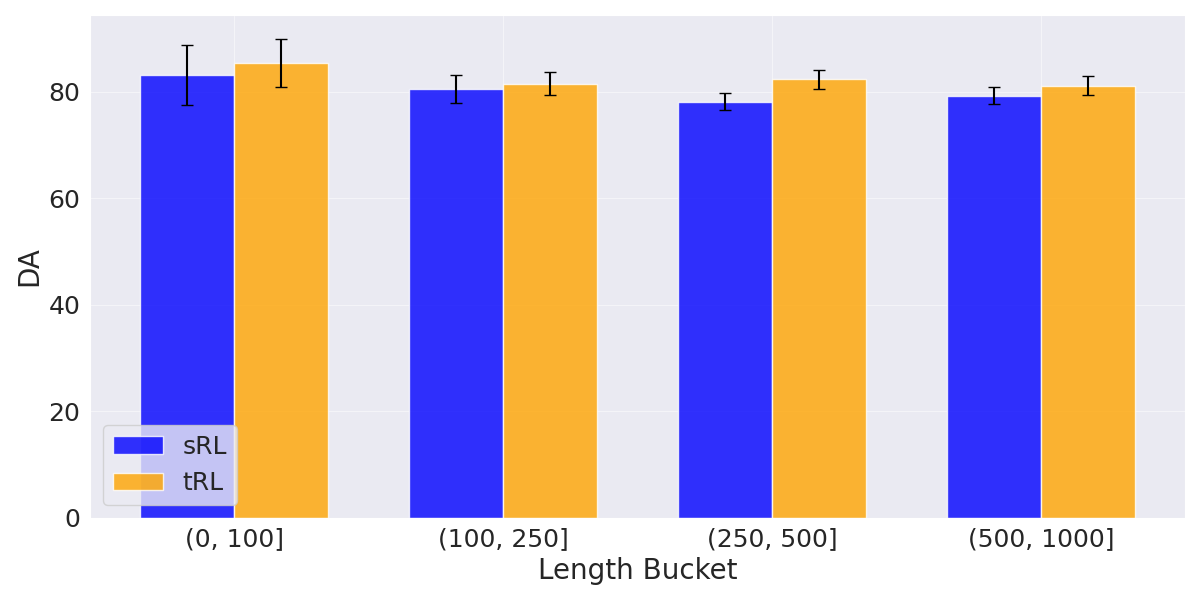}
    \caption{DA scores for sRL and tRL with \textsc{Tower} on WMT24 EN$\to$DE across increasing source sentence lengths.}
    \label{fig:human_eval_sent_wmt24}
\end{figure}

\paragraph{Findings.}
As shown in Table~\ref{tab:human_eval}, both sRL and tRL models consistently outperform SFT and CPO, demonstrating their advantage in translation quality. On the WMT24 dataset, tRL achieves an average DA score of $82.6$, $2.3$ points higher than sRL, with consistent gains across sentence-length buckets, as shown in Figure~\ref{fig:human_eval_sent_wmt24}; longer sentences exhibit more significant improvements based on error bar overlap. Human evaluations were conducted by two professional annotators, and inter-annotator agreement is moderate-to-high (Pearson's $r=0.59$, Spearman's $\rho=0.57$), indicating reliable scoring. 
These results align with our automatic evaluation, including the ablation analysis, which collectively shows that tRL enhances stability and translation quality, particularly for longer sentences.

%% file: tables/main_results.tex
\begin{table*}[!ht]
    \centering
    \scriptsize
    {\renewcommand{\arraystretch}{1.02}
    \resizebox{0.9\linewidth}{!}{
    \setlength\tabcolsep{3pt}
    \begin{tabular}{lcccccc}\toprule
        \multirow{2}{*}{{\textsc{Model}}} & \multicolumn{6}{c}{Metrics} \\ \cmidrule(lr){2-7}
        & \textsc{BLEU} & \textsc{ChrF}  & \textsc{COMET22} & \xcomet & \textsc{BLEURT} & \textsc{CometKiwi-23} \\
        \midrule
        & \multicolumn{6}{c}{WMT18 EN→DE} \\ 
        \midrule
         NLLB & 41.60\seventhcluster & 66.34\fourthcluster & 86.94\sixthcluster & 94.69\fourthcluster & 76.42\fourthcluster & 72.46\thirdcluster \\
         ~\quad + SFT & 43.07\secondcluster & 67.38\secondcluster & 87.18\thirdcluster & 94.97\fourthcluster & 76.35\fourthcluster & 72.57\thirdcluster \\
        ~\quad ~\quad + sRL w/ \textsc{BLEU} & \textbf{43.31}\firstcluster & \textbf{67.57}\firstcluster & 87.31\secondcluster & 95.13\thirdcluster & 76.61\thirdcluster & 72.73\secondcluster \\
         ~\quad ~\quad + sRL w/ \xcomet & 43.09\secondcluster & 67.34\secondcluster & 87.18\thirdcluster & 95.55\secondcluster & 76.72\secondcluster & 72.84\secondcluster \\
         ~\quad ~\quad + sRL w/ \xcometmqm & 43.03\thirdcluster & 67.29\thirdcluster & 87.10\fourthcluster & 95.50\secondcluster & 76.72\secondcluster & 72.40\thirdcluster \\
       ~\quad ~\quad + tRL w/ \textsc{BLEU} & 42.90\fourthcluster & 67.32\thirdcluster & 87.15\thirdcluster & 94.94\fourthcluster & 76.37\fourthcluster & 72.57\thirdcluster \\
        ~\quad ~\quad + tRL w/ \xcometmqm & 42.63\fifthcluster & 67.47\secondcluster & \textbf{88.27}\firstcluster & \textbf{96.18}\firstcluster & \textbf{77.60}\firstcluster & \textbf{75.16}\firstcluster \\
        ~\quad + CPO w/ \xcomet & 42.30\sixthcluster & 66.10\fourthcluster & 87.02\fifthcluster & 95.31\thirdcluster & 76.70\secondcluster & 73.08\secondcluster \\
        \midrule
        & \multicolumn{6}{c}{WMT18 DE→EN} \\            
        \midrule
        NLLB & 43.50\fifthcluster & 66.12\fifthcluster & 86.45\fifthcluster & 93.95\fourthcluster & 75.84\fourthcluster & 73.16\fifthcluster \\
        ~\quad + SFT & 45.57\secondcluster & 67.83\secondcluster & 87.22\thirdcluster & 95.08\thirdcluster & 76.79\thirdcluster & 74.11\thirdcluster \\        
        ~\quad ~\quad + sRL w/ \textsc{BLEU} & \textbf{45.78}\firstcluster & \textbf{67.97}\firstcluster & 87.26\secondcluster & 95.07\thirdcluster & 76.85\secondcluster & 74.05\thirdcluster \\
        ~\quad ~\quad + sRL w/ \xcomet & 45.51\thirdcluster & 67.78\thirdcluster & 87.21\thirdcluster & 95.28\secondcluster & 76.76\thirdcluster & 74.06\thirdcluster \\
        ~\quad ~\quad + sRL w/ \xcometmqm & 45.55\thirdcluster & 67.80\thirdcluster & 87.21\thirdcluster & 95.17\thirdcluster & 76.80\thirdcluster & 74.11\thirdcluster \\
       ~\quad ~\quad + tRL w/ \textsc{BLEU} & 45.53\thirdcluster & 67.74\thirdcluster & 87.29\secondcluster & 95.14\thirdcluster & 76.84\secondcluster & 74.00\fourthcluster \\
       ~\quad ~\quad + tRL w/ \xcometmqm & 45.71\secondcluster & 67.60\fourthcluster & \textbf{87.91}\firstcluster & \textbf{96.02}\firstcluster & \textbf{77.70}\firstcluster & \textbf{74.42}\firstcluster \\
        ~\quad + CPO w/ \xcomet & 43.88\fourthcluster & 66.05\fifthcluster & 87.10\fourthcluster & 95.33\secondcluster & 77.08\secondcluster & 74.23\secondcluster \\
        \midrule
        & \multicolumn{6}{c}{IWSLT2017 EN→FR} \\            
        \midrule
        NLLB & 43.61\fourthcluster & 65.76\thirdcluster & 84.93\thirdcluster & 90.27\thirdcluster & 72.61\fourthcluster & 70.06\fourthcluster \\
        ~\quad + SFT & 45.63\secondcluster & 67.59\secondcluster & 86.01\secondcluster & 91.07\secondcluster & 74.41\secondcluster & 71.78\secondcluster \\
        ~\quad ~\quad + sRL w/ \textsc{BLEU} & 45.77\secondcluster & \textbf{67.68\firstcluster} & 86.04\secondcluster & 91.11\secondcluster & 74.44\secondcluster & 71.79\secondcluster \\
        ~\quad ~\quad + sRL w/ \xcomet & 45.68\secondcluster & 67.61\secondcluster & 86.03\secondcluster & 91.67\secondcluster & 74.45\secondcluster & 71.77\secondcluster \\
        ~\quad ~\quad + sRL w/ \xcometmqm & 45.76\secondcluster & 67.60\secondcluster & 86.08\secondcluster & 91.68\secondcluster & 74.40\secondcluster & 71.77\secondcluster \\
        ~\quad ~\quad + tRL w/ \textsc{BLEU} & 45.71\secondcluster & 67.63\secondcluster & 86.00\secondcluster & 91.14\secondcluster & 74.48\secondcluster & 71.82\secondcluster \\
        ~\quad ~\quad + tRL w/ \xcometmqm & \textbf{46.58\firstcluster} & 60.07\fourthcluster & \textbf{87.17\firstcluster} & \textbf{92.17\firstcluster} & \textbf{75.58\firstcluster} & \textbf{72.80\firstcluster} \\
        ~\quad + CPO w/ \xcomet & 43.96\thirdcluster & 65.24\thirdcluster & 85.55\thirdcluster & 91.65\secondcluster & 74.02\thirdcluster & 71.86\thirdcluster \\
        \midrule
        & \multicolumn{6}{c}{IWSLT2017 FR→EN} \\            
        \midrule
        NLLB & 45.76\thirdcluster & 65.76\thirdcluster & 87.15\thirdcluster & 94.69\thirdcluster & 77.55\thirdcluster & 71.99\thirdcluster \\
        ~\quad + SFT & 48.65\secondcluster & 67.69\secondcluster & 88.22\secondcluster & 95.34\secondcluster & 78.74\secondcluster & 73.59\secondcluster \\
        ~\quad ~\quad + sRL w/ \textsc{BLEU} & 48.76\secondcluster & 67.76\secondcluster & 88.22\secondcluster & 95.33\secondcluster & 78.74\secondcluster & 73.54\secondcluster \\
        ~\quad ~\quad + sRL w/ \xcomet & 48.62\secondcluster & \textbf{67.88\firstcluster} & 88.30\secondcluster & \textbf{95.53\firstcluster} & 78.73\secondcluster & 73.56\secondcluster \\
        ~\quad ~\quad + sRL w/ \xcometmqm & 48.61\secondcluster & 67.66\secondcluster & 88.31\secondcluster & \textbf{95.53\firstcluster} & 78.74\secondcluster & 73.56\secondcluster \\
        ~\quad ~\quad + tRL w/ \textsc{BLEU} & 48.56\secondcluster & 67.69\secondcluster & 88.21\secondcluster & 95.33\secondcluster & 78.73\secondcluster & 73.60\secondcluster \\
        ~\quad ~\quad + tRL w/ \xcometmqm & \textbf{49.06\firstcluster} & 61.17\fourthcluster & \textbf{88.46\firstcluster} & 95.27\secondcluster & \textbf{79.10\firstcluster} & \textbf{74.74\firstcluster} \\
        ~\quad + CPO w/ \xcomet & 48.46\secondcluster & 67.54\secondcluster & 88.20\secondcluster & 95.42\secondcluster & 78.71\secondcluster & 73.52\secondcluster \\
        
        \bottomrule
    \end{tabular}
    }
    \caption{\label{tab:results} Evaluation of NLLB models on WMT18 (EN$\leftrightarrow$DE) and IWSLT2017 (EN$\leftrightarrow$FR), with rows grouped by test set. We provide automatic evaluation metrics for the best base model, baseline (fine-tuned base model) in each dataset and the variations with sentence-level and token-level RL training. \textsc{BLEU} and \xcomet serve as reward models in the context of RL training. MQM scores are predicted from the error spans ($y = y_\mathrm{MQM}$) \cite{xcomet}. 
    Best-performing values are \textbf{bolded}, and models are grouped into statistically significant quality clusters.}
    }
    \vspace{-1em}
\end{table*}

%% file: tables/llm_results.tex
\begin{table*}[!ht]
   \centering
    \scriptsize	
    {\renewcommand{\arraystretch}{1.02}
    \resizebox{0.9\linewidth}{!}{
    \setlength\tabcolsep{3pt}
    \begin{tabular}{lcccccc}\toprule
\multirow{2}{*}{{\textsc{Model}}} & \multicolumn{6}{c}{Metrics} \\ \cmidrule(lr){2-7}
        & \textsc{BLEU} & \textsc{ChrF}  & \textsc{COMET22} & \xcomet & \textsc{BLEURT} & \textsc{CometKiwi-23} \\
        \midrule
        & \multicolumn{6}{c}{WMT24 EN→DE} \\ 
        \midrule
        \textsc{NLLB} & 35.01\sixthcluster & 53.92\fifthcluster & 64.30\fourthcluster & 83.40\secondcluster & 60.90\thirdcluster & 55.50\sixthcluster \\
        ~\quad + SFT & 37.92\thirdcluster & 59.29\thirdcluster & 65.20\secondcluster & 82.50\fifthcluster & 61.41\secondcluster & 58.20\thirdcluster \\
        ~\quad ~\quad + sRL w/ \textsc{BLEU} & 38.54\secondcluster & 59.77\secondcluster & 65.50\secondcluster & 82.50\fifthcluster & 61.40\secondcluster & 58.30\thirdcluster \\
        ~\quad ~\quad + sRL w/ \xcomet & 37.88\thirdcluster & 59.16\thirdcluster & 65.50\secondcluster & 82.80\fourthcluster & 60.40\fourthcluster & 57.90\fourthcluster \\
        ~\quad ~\quad + sRL w/ \xcometmqm & 37.14\fourthcluster & 59.29\thirdcluster & 65.40\secondcluster & 83.30\secondcluster & 60.40\fourthcluster & 57.50\fifthcluster \\
        ~\quad ~\quad + tRL w/ \textsc{BLEU} & 38.53\secondcluster & 59.60\secondcluster & 65.49\secondcluster & 82.80\fourthcluster & 61.41\secondcluster & 59.10\secondcluster \\
        ~\quad ~\quad + tRL w/ \xcometmqm & \textbf{39.42}\firstcluster & \textbf{60.40}\firstcluster & \textbf{66.43}\firstcluster & \textbf{83.60}\firstcluster & \textbf{62.29}\firstcluster & \textbf{60.70}\firstcluster \\
         ~\quad + CPO w/ \xcomet & 36.07\fifthcluster & 55.32\fourthcluster & 64.70\thirdcluster & 82.90\fourthcluster & 60.80\thirdcluster & 58.10\thirdcluster \\ \midrule
        \textsc{Tower} & 42.77\sixthcluster & 62.60\fifthcluster & 71.80\fifthcluster & 88.50\fourthcluster & 68.59\fifthcluster & 67.20\fourthcluster \\
        ~\quad + SFT & 45.14\fourthcluster & 63.30\fourthcluster & 72.18\fourthcluster & 88.90\thirdcluster & 69.18\fourthcluster & 67.30\fourthcluster  \\
        ~\quad ~\quad + sRL w/ \textsc{BLEU} & 46.07\secondcluster & 63.77\thirdcluster & 72.26\fourthcluster & 89.20\thirdcluster & 70.01\thirdcluster & 68.20\thirdcluster \\
        ~\quad ~\quad + sRL w/ \xcomet & 45.59\thirdcluster & 64.11\secondcluster & 72.73\secondcluster & 90.50\secondcluster & 70.58\secondcluster & 69.90\secondcluster  \\
        ~\quad ~\quad + sRL w/ \xcometmqm & 45.58\thirdcluster & 63.76\thirdcluster & 72.41\thirdcluster & 90.20\secondcluster & 70.55\secondcluster & 69.70\secondcluster \\
        ~\quad ~\quad + tRL w/ \textsc{BLEU} & 46.13\secondcluster & 64.15\secondcluster & 72.22\fourthcluster & 89.22\thirdcluster & 70.11\thirdcluster & 68.33\thirdcluster \\
        ~\quad ~\quad + tRL w/ \xcometmqm & \textbf{46.92}\firstcluster & \textbf{65.63}\firstcluster & \textbf{74.66}\firstcluster & \textbf{91.90}\firstcluster & \textbf{71.80}\firstcluster & \textbf{71.20}\firstcluster  \\
         ~\quad + CPO w/ \xcomet & 43.15\fifthcluster & 61.13\sixthcluster & 70.65\sixthcluster & 87.80\fifthcluster &  68.23\fifthcluster & 67.38\fourthcluster \\
        \midrule
        \textsc{Gemma} & 15.13\seventhcluster & 51.99\fifthcluster & 54.84\seventhcluster & 61.40\fifthcluster & 51.35\sixthcluster & 42.80\sixthcluster \\
        ~\quad + SFT & 35.19\fifthcluster & 59.03\thirdcluster & 69.86\fifthcluster & 86.10\fourthcluster & 66.13\fifthcluster & 64.30\fifthcluster  \\
        ~\quad ~\quad + sRL w/ \textsc{BLEU} & 36.34\thirdcluster & 59.33\secondcluster & 69.96\fifthcluster & 86.10\fourthcluster & 66.50\fourthcluster & 64.40\fifthcluster \\
        ~\quad ~\quad + sRL w/ \xcomet & 36.16\thirdcluster & 59.28\secondcluster & 70.82\secondcluster & 87.20\secondcluster & 66.90\thirdcluster & 66.30\secondcluster \\
        ~\quad ~\quad + sRL w/ \xcometmqm & 35.98\fourthcluster & 59.33\secondcluster & 70.39\thirdcluster & 87.10\secondcluster & 67.20\secondcluster & 65.70\thirdcluster \\
        ~\quad ~\quad + tRL w/ \textsc{BLEU} & 36.63\secondcluster & 59.52\firstcluster & 70.02\fourthcluster & 86.40\thirdcluster & 66.70\thirdcluster & 64.50\fifthcluster \\
        ~\quad ~\quad + tRL w/ \xcometmqm & \textbf{36.90}\firstcluster & \textbf{59.58}\firstcluster & \textbf{71.12}\firstcluster & \textbf{88.10}\firstcluster & \textbf{68.00}\firstcluster & \textbf{66.90}\firstcluster \\
         ~\quad + CPO w/ \xcomet & 35.05\sixthcluster & 58.52\fourthcluster & 69.22\sixthcluster & 86.40\thirdcluster & 66.80\thirdcluster & 65.10\fourthcluster  \\
        \bottomrule
    \end{tabular}}
    \caption{Evaluation metrics for \textsc{NLLB}, \textsc{Tower}, \textsc{Gemma} and its variations across WMT24 EN→DE. Best-performing values are \textbf{bolded}, and models are grouped into statistically significant quality clusters.}
    \label{tab:results_wmt24}}
    \vspace{-1em}
\end{table*}

%% file: tables/severity_map.tex
\begin{table}[!ht]
\centering
\footnotesize
\setlength{\tabcolsep}{4pt}
\scalebox{0.9}{
\begin{tabular}{lccccc}
\toprule
\textsc{\textbf{Word}} & \textsc{\textbf{Bin}} & \textsc{\textbf{MQM}} & \textsc{\textbf{rMQM}} & \textsc{\textbf{Our}} & \textsc{\textbf{rOur}} \\ 
\midrule
\textsc{\textbf{Correct}} & 1 & 0 & 25 & 8 & -1 \\ 
\textsc{\textbf{Minor}}  & -1  & -1 & 5 & 4 & -2 \\ 
\textsc{\textbf{Major}} & -1  & -5  & 1 & 2 & -4 \\ 
\textsc{\textbf{Critical}} & -1 & -25 & 0 & 1 & -8 \\ 
\bottomrule
\end{tabular}}
\caption{Severity maps.}
\label{tab:table_severity_map}
\vspace{-1em}
\end{table}

%% file: tables/results_severity_map.tex
\begin{table*}[!ht]
    \centering
    \scriptsize	
    {\renewcommand{\arraystretch}{1.02}
    \resizebox{\linewidth}{!}{
    \setlength\tabcolsep{3pt}
    \begin{tabular}{lccccccc}\toprule
        \multirow{2.5}{*}{{\textsc{Model}}} & \multirow{2.5}{*}{{\textsc{Severity Map}}} & \multicolumn{6}{c}{Metrics}\\ \cmidrule(lr){3-8}
        & & \textsc{BLEU} & \textsc{ChrF}  & \textsc{COMET22} & \textsc{xCOMET} & \textsc{BLEURT} & \textsc{CometKiwi-23} \\
        \midrule
        & \multicolumn{6}{c}{IWSLT2017 EN→FR} \\ 
        \midrule        
        \textsc{NLLB} & -- & 43.61 & 65.76 & 84.93 & 90.27 & 72.61 & 70.06  \\
        ~\quad + SFT & -- & 45.63 & \textbf{67.59} & 86.01 & 91.07 & 74.41 & 71.78 \\
       ~\quad ~\quad + tRL w/ \xcometmqm & \textsc{Bin} & 45.44 & 60.43 & 85.87 & 90.90 & 74.23 & 71.58 \\
       ~\quad ~\quad + tRL w/ \xcometmqm & \textsc{MQM} & 45.68 & 60.67 & 85.97 & 91.01 & 74.33 & 71.74 \\
       ~\quad ~\quad + tRL w/ \xcometmqm & \textsc{rMQM} & 41.47 & 59.74 & 82.20 & 84.58 & 70.45 & 67.29 \\
       ~\quad ~\quad + tRL w/ \xcometmqm & \textsc{Our} & \textbf{46.58} & 60.07 & \textbf{87.17} & \textbf{92.17} & \textbf{75.58} & \textbf{72.80} \\
       ~\quad ~\quad + tRL w/ \xcometmqm & \textsc{rOur} & 45.89 & 60.10 & 85.90 & 91.33 & 74.39 & 71.79 \\
        \bottomrule
    \end{tabular}}
     \caption{Automatic evaluation metrics for several severity maps setup in the context of token-level RL training. Best-performing values are \textbf{bolded}.}
    \label{tab:results_severity_map}}
    \vspace{-1em}
\end{table*}

%% file: tables/results_rl_method.tex
\begin{table*}[!ht]
    \centering
    \scriptsize	
    {\renewcommand{\arraystretch}{1.02}
    \resizebox{\linewidth}{!}{
    \setlength\tabcolsep{4pt}
    \begin{tabular}{lcccccc}\toprule
        \multirow{2}{*}{{\textsc{Model}}} & \multicolumn{6}{c}{Metrics}\\ \cmidrule(lr){2-7}
        & \textsc{BLEU} & \textsc{ChrF}  & \textsc{COMET22} & \xcomet & \textsc{BLEURT} & \textsc{CometKiwi-23} \\
        \midrule
        & \multicolumn{6}{c}{IWSLT2017 EN→FR} \\ 
        \midrule  
        \textsc{NLLB} & 43.61 & 65.76 & 84.93 & 90.27 & 72.61 & 70.06  \\
        ~\quad + SFT & 45.63 & 67.59 & 86.01 & 91.07 & 74.41 & 71.78 \\
        ~\quad ~\quad + sREINFORCE w/ \xcomet  & 45.70 & 67.58 & 86.11 & 91.29 & 74.43 & 71.75 \\
        ~\quad ~\quad + sREINFORCE w/ \xcometmqm  & 45.72  & 67.59 & 86.07 & 91.35 & 74.78 & 71.80 \\
        ~\quad ~\quad + tREINFORCE w/ \xcometmqm  & 46.07 & 60.60 & 86.24  & 91.77 & 74.98 & 71.87 \\
        ~\quad ~\quad + sPPO w/ \xcomet & 45.68 & \textbf{67.61} & 86.03 & 91.67 & 74.45 & 71.77 \\
        ~\quad ~\quad + sPPO w/ \xcometmqm & 45.76 & 67.60 & 86.08 & 91.68 & 74.40 & 71.77 \\
        ~\quad ~\quad + tPPO w/ \xcometmqm & \textbf{46.58} & 60.07 & \textbf{87.17} & \textbf{92.17} & \textbf{75.58} & \textbf{72.80} \\
        \bottomrule
    \end{tabular}}
    \caption{Automatic evaluation metrics for REINFORCE and PPO in the context of token-level RL training. Best-performing values are \textbf{bolded}.}
    \label{tab:results_rl_method}}
    \vspace{-1em}
\end{table*}

%% file: tables/human_eval.tex
\begin{table}[!ht]
\centering
\footnotesize
\scalebox{1}{
\begin{tabular}{lccccc}
\toprule
\textsc{\textbf{}} & \textsc{\textbf{CPO}} & \textsc{\textbf{SFT}} & \textsc{\textbf{tRL}} & \textsc{\textbf{sRL}} & \textsc{\textbf{Reference}} \\ 
\midrule
\textsc{\textbf{Ann1}} & 59.5 & 64.5 & 66.9 & 66.8 & 76.0 \\ 
\textsc{\textbf{Ann2}} & 53.2 & 54.2 & 56.1 & 57.6 & 60.0 \\ 
\bottomrule
\end{tabular}}
\caption{Human Evaluation on WMT18.}
\label{tab:human_eval}
\end{table}

%% file: conclusion.tex
\section{Conclusion}
In this work, we propose a new method for improving NMT that uses fine-grained reward optimization with \xcomet as a token-level reward model. 
While exposure bias arises from SFT, sentence-level RL addresses this issue but introduces reward sparsity due to coarse-grained feedback. 
Our token-level RL approach overcomes this by providing a denser and more informative reward signal to enhance translation quality.
Our experiments show that incorporating fine-grained reward mechanisms significantly improves MT quality, especially for longer sequences, and also stabilizes training. 
Additionally, token-level RL training outperforms sentence-level RL training in most evaluation metrics.
Our findings show that fine-grained RL offers a more effective MT optimization framework by mitigating reward sparsity and aligning better with human judgments.

%% file: bibliography.bib
@inproceedings{BENGIO2015,
	title        = {Scheduled Sampling for Sequence Prediction with Recurrent Neural Networks},
	author       = {Bengio, Samy and Vinyals, Oriol and Jaitly, Navdeep and Shazeer, Noam},
	year         = 2015,
	booktitle    = {Advances in Neural Information Processing Systems},
	publisher    = {Curran Associates, Inc.},
	volume       = 28,
	pages        = {},
	url          = {https://proceedings.neurips.cc/paper_files/paper/2015/file/e995f98d56967d946471af29d7bf99f1-Paper.pdf},
	editor       = {C. Cortes and N. Lawrence and D. Lee and M. Sugiyama and R. Garnett}
}

@inproceedings{almapref,
	title        = {A Paradigm Shift in Machine Translation: Boosting Translation Performance of Large Language Models},
	author       = {Haoran Xu and Young Jin Kim and Amr Sharaf and Hany Hassan Awadalla},
	year         = 2024,
	booktitle    = {ICLR},
	url          = {https://openreview.net/forum?id=farT6XXntP},
	cdate        = 1704067200000
}

@inproceedings{he-etal-2024-improving,
	title        = {Improving Machine Translation with Human Feedback: An Exploration of Quality Estimation as a Reward Model},
	author       = {He, Zhiwei  and Wang, Xing  and Jiao, Wenxiang  and Zhang, Zhuosheng  and Wang, Rui  and Shi, Shuming  and Tu, Zhaopeng},
	year         = 2024,
	month        = jun,
	booktitle    = {Proceedings of the 2024 Conference of the North American Chapter of the Association for Computational Linguistics: Human Language Technologies (Volume 1: Long Papers)},
	publisher    = {Association for Computational Linguistics},
	address      = {Mexico City, Mexico},
	pages        = {8164--8180},
	doi          = {10.18653/v1/2024.naacl-long.451},
	url          = {https://aclanthology.org/2024.naacl-long.451/},
	editor       = {Duh, Kevin  and Gomez, Helena  and Bethard, Steven}
}

@inproceedings{Zhu2024,
	title        = {A Preference-driven Paradigm for Enhanced Translation with Large Language Models},
	author       = {Zhu, Dawei  and Trenous, Sony  and Shen, Xiaoyu  and Klakow, Dietrich  and Byrne, Bill  and Hasler, Eva},
	year         = 2024,
	month        = jun,
	booktitle    = {Proceedings of the 2024 Conference of the North American Chapter of the Association for Computational Linguistics: Human Language Technologies (Volume 1: Long Papers)},
	publisher    = {Association for Computational Linguistics},
	address      = {Mexico City, Mexico},
	pages        = {3385--3403},
	doi          = {10.18653/v1/2024.naacl-long.186},
	url          = {https://aclanthology.org/2024.naacl-long.186/},
	editor       = {Duh, Kevin  and Gomez, Helena  and Bethard, Steven}
}

@inproceedings{agrawal-etal-2024-modeling,
	title        = {Modeling User Preferences with Automatic Metrics: Creating a High-Quality Preference Dataset for Machine Translation},
	author       = {Agrawal, Sweta  and De Souza, Jos{\'e} G. C.  and Rei, Ricardo  and Farinhas, Ant{\'o}nio  and Faria, Gon{\c{c}}alo  and Fernandes, Patrick  and Guerreiro, Nuno M  and Martins, Andre},
	year         = 2024,
	month        = nov,
	booktitle    = {Proceedings of the 2024 Conference on Empirical Methods in Natural Language Processing},
	publisher    = {Association for Computational Linguistics},
	address      = {Miami, Florida, USA},
	pages        = {14503--14519},
	doi          = {10.18653/v1/2024.emnlp-main.803},
	url          = {https://aclanthology.org/2024.emnlp-main.803/},
	editor       = {Al-Onaizan, Yaser  and Bansal, Mohit  and Chen, Yun-Nung}
}

@article{domingo2017segment,
	title        = {Segment-based interactive-predictive machine translation},
	author       = {Domingo, Miguel and Peris, \'{A}lvaro and Casacuberta, Francisco},
	year         = 2017,
	month        = dec,
	journal      = {Machine Translation},
	publisher    = {Kluwer Academic Publishers},
	address      = {USA},
	volume       = 31,
	number       = 4,
	pages        = {163–185},
	doi          = {10.1007/s10590-017-9213-3},
	issn         = {0922-6567},
	url          = {https://doi.org/10.1007/s10590-017-9213-3},
	issue_date   = {December 2017},
	numpages     = 23
}

@article{lommel-mqm,
	title        = {Multidimensional Quality Metrics ({MQM}): A Framework for Declaring and Describing Translation Quality Metrics},
	author       = {Lommel, Arle and Burchardt, Aljoscha and Uszkoreit, Hans},
	year         = 2014,
	month        = 12,
	journal      = {Tradumàtica: tecnologies de la traducció},
	volume       = {0},
	pages        = {455--463},
	doi          = {10.5565/rev/tradumatica.77}
}

@incollection{PUTERMAN1990,
	title        = {Chapter 8 Markov decision processes},
	author       = {Martin L. Puterman},
	year         = 1990,
	booktitle    = {Stochastic Models},
	publisher    = {Elsevier},
	series       = {Handbooks in Operations Research and Management Science},
	volume       = 2,
	pages        = {331--434},
	doi          = {https://doi.org/10.1016/S0927-0507(05)80172-0},
	issn         = {0927-0507},
	url          = {https://www.sciencedirect.com/science/article/pii/S0927050705801720}
}

@article{WILLIAMS1992,
	title        = {Simple statistical gradient-following algorithms for connectionist reinforcement learning},
	author       = {Williams, Ronald J.},
	year         = 1992,
	month        = {May},
	day          = {01},
	journal      = {Machine Learning},
	volume       = 8,
	number       = 3,
	pages        = {229--256},
	doi          = {10.1007/BF00992696},
	issn         = {1573-0565},
	url          = {https://doi.org/10.1007/BF00992696}
}

@article{wu2016google,
	title        = {{Google's Neural Machine Translation System: Bridging the Gap between Human and Machine Translation}},
	author       = {{Wu}, Yonghui and {Schuster}, Mike and {Chen}, Zhifeng and {Le}, Quoc V. and {Norouzi}, Mohammad and {Macherey}, Wolfgang and {Krikun}, Maxim and {Cao}, Yuan and {Gao}, Qin and {Macherey}, Klaus and {Klingner}, Jeff and {Shah}, Apurva and {Johnson}, Melvin and {Liu}, Xiaobing and {Kaiser}, {\L}ukasz and {Gouws}, Stephan and {Kato}, Yoshikiyo and {Kudo}, Taku and {Kazawa}, Hideto and {Stevens}, Keith and {Kurian}, George and {Patil}, Nishant and {Wang}, Wei and {Young}, Cliff and {Smith}, Jason and {Riesa}, Jason and {Rudnick}, Alex and {Vinyals}, Oriol and {Corrado}, Greg and {Hughes}, Macduff and {Dean}, Jeffrey},
	year         = 2016,
	month        = sep,
	journal      = {arXiv e-prints},
	pages        = {arXiv:1609.08144},
	doi          = {10.48550/arXiv.1609.08144},
	eid          = {arXiv:1609.08144},
	archiveprefix = {arXiv},
	eprint       = {1609.08144},
	primaryclass = {cs.CL},
	adsurl       = {https://ui.adsabs.harvard.edu/abs/2016arXiv160908144W}
}

@book{sutton2018reinforcement,
	title        = {Reinforcement learning: An introduction, 2nd ed.},
	author       = {Sutton, Richard S. and Barto, Andrew G.},
	year         = 2018,
	publisher    = {The MIT Press},
	address      = {Cambridge,  MA,  US},
	series       = {Adaptive computation and machine learning.},
	pages        = {xxii, 526-xxii, 526},
	isbn         = {9780262039246 (Hardcover)}
}

@inproceedings{wmt18,
	title        = {Findings of the 2018 Conference on Machine Translation ({WMT}18)},
	author       = {Bojar, Ond{\v{r}}ej  and Federmann, Christian  and Fishel, Mark  and Graham, Yvette  and Haddow, Barry  and Huck, Matthias  and Koehn, Philipp  and Monz, Christof},
	year         = 2018,
	month        = oct,
	booktitle    = {Proceedings of the Third Conference on Machine Translation: Shared Task Papers},
	publisher    = {Association for Computational Linguistics},
	address      = {Belgium, Brussels},
	pages        = {272--303},
	doi          = {10.18653/v1/W18-6401},
	url          = {https://aclanthology.org/W18-6401},
	editor       = {Bojar, Ond{\v{r}}ej  and Chatterjee, Rajen  and Federmann, Christian  and Fishel, Mark  and Graham, Yvette  and Haddow, Barry  and Huck, Matthias  and Yepes, Antonio Jimeno  and Koehn, Philipp  and Monz, Christof  and Negri, Matteo  and N{\'e}v{\'e}ol, Aur{\'e}lie  and Neves, Mariana  and Post, Matt  and Specia, Lucia  and Turchi, Marco  and Verspoor, Karin}
}

@inproceedings{iwslt2017,
	title        = {Overview of the {IWSLT} 2017 Evaluation Campaign},
	author       = {Cettolo, Mauro  and Federico, Marcello  and Bentivogli, Luisa  and Niehues, Jan  and St{\"u}ker, Sebastian  and Sudoh, Katsuhito  and Yoshino, Koichiro  and Federmann, Christian},
	year         = 2017,
	month        = {dec} # { 14-15},
	booktitle    = {Proceedings of the 14th International Conference on Spoken Language Translation},
	publisher    = {International Workshop on Spoken Language Translation},
	address      = {Tokyo, Japan},
	pages        = {2--14},
	url          = {https://aclanthology.org/2017.iwslt-1.1},
	editor       = {Sakti, Sakriani  and Utiyama, Masao}
}

@inproceedings{Kalchbrenner2013RecurrentCT,
	title        = {Recurrent Continuous Translation Models},
	author       = {Kalchbrenner, Nal  and Blunsom, Phil},
	year         = 2013,
	month        = oct,
	booktitle    = {Proceedings of the 2013 Conference on Empirical Methods in Natural Language Processing},
	publisher    = {Association for Computational Linguistics},
	address      = {Seattle, Washington, USA},
	pages        = {1700--1709},
	url          = {https://aclanthology.org/D13-1176},
	editor       = {Yarowsky, David  and Baldwin, Timothy  and Korhonen, Anna  and Livescu, Karen  and Bethard, Steven}
}

@inproceedings{ng1999policy,
	title        = {Policy Invariance Under Reward Transformations: Theory and Application to Reward Shaping},
	author       = {Ng, Andrew Y. and Harada, Daishi and Russell, Stuart J.},
	year         = 1999,
	booktitle    = {Proceedings of the Sixteenth International Conference on Machine Learning},
	publisher    = {Morgan Kaufmann Publishers Inc.},
	address      = {San Francisco, CA, USA},
	series       = {ICML '99},
	pages        = {278–287},
	isbn         = 1558606122,
	numpages     = 10
}

@inproceedings{bleu,
	title        = {BLEU: a method for automatic evaluation of machine translation},
	author       = {Papineni, Kishore and Roukos, Salim and Ward, Todd and Zhu, Wei-Jing},
	year         = 2002,
	booktitle    = {Proceedings of the 40th Annual Meeting on Association for Computational Linguistics},
	location     = {Philadelphia, Pennsylvania},
	publisher    = {Association for Computational Linguistics},
	address      = {USA},
	series       = {ACL '02},
	pages        = {311–318},
	doi          = {10.3115/1073083.1073135},
	url          = {https://doi.org/10.3115/1073083.1073135},
	numpages     = 8
}

@inproceedings{chrf,
	title        = {chr{F}: character n-gram {F}-score for automatic {MT} evaluation},
	author       = {Popovi{\'c}, Maja},
	year         = 2015,
	month        = sep,
	booktitle    = {Proceedings of the Tenth Workshop on Statistical Machine Translation},
	publisher    = {Association for Computational Linguistics},
	address      = {Lisbon, Portugal},
	pages        = {392--395},
	doi          = {10.18653/v1/W15-3049},
	url          = {https://aclanthology.org/W15-3049},
	editor       = {Bojar, Ond{\v{r}}ej  and Chatterjee, Rajan  and Federmann, Christian  and Haddow, Barry  and Hokamp, Chris  and Huck, Matthias  and Logacheva, Varvara  and Pecina, Pavel}
}

@inproceedings{RANZATO2016,
	title        = {Sequence Level Training with Recurrent Neural Networks},
	author       = {Marc'Aurelio Ranzato and Sumit Chopra and Michael Auli and Wojciech Zaremba},
	year         = 2016,
	booktitle    = {4th International Conference on Learning Representations, {ICLR} 2016, San Juan, Puerto Rico, May 2-4, 2016, Conference Track Proceedings},
	url          = {http://arxiv.org/abs/1511.06732},
	editor       = {Yoshua Bengio and Yann LeCun},
	bibsource    = {dblp computer science bibliography, https://dblp.org}
}

@inproceedings{comet22,
	title        = {{COMET}-22: Unbabel-{IST} 2022 Submission for the Metrics Shared Task},
	author       = {Rei, Ricardo  and C. de Souza, Jos{\'e} G.  and Alves, Duarte  and Zerva, Chrysoula  and Farinha, Ana C  and Glushkova, Taisiya  and Lavie, Alon  and Coheur, Luisa  and Martins, Andr{\'e} F. T.},
	year         = 2022,
	month        = dec,
	booktitle    = {Proceedings of the Seventh Conference on Machine Translation (WMT)},
	publisher    = {Association for Computational Linguistics},
	address      = {Abu Dhabi, United Arab Emirates (Hybrid)},
	pages        = {578--585},
	url          = {https://aclanthology.org/2022.wmt-1.52},
	editor       = {Koehn, Philipp  and Barrault, Lo{\"\i}c  and Bojar, Ond{\v{r}}ej  and Bougares, Fethi  and Chatterjee, Rajen  and Costa-juss{\`a}, Marta R.  and Federmann, Christian  and Fishel, Mark  and Fraser, Alexander  and Freitag, Markus  and Graham, Yvette  and Grundkiewicz, Roman  and Guzman, Paco  and Haddow, Barry  and Huck, Matthias  and Jimeno Yepes, Antonio  and Kocmi, Tom  and Martins, Andr{\'e}  and Morishita, Makoto  and Monz, Christof  and Nagata, Masaaki  and Nakazawa, Toshiaki  and Negri, Matteo  and N{\'e}v{\'e}ol, Aur{\'e}lie  and Neves, Mariana  and Popel, Martin  and Turchi, Marco  and Zampieri, Marcos}
}

@inproceedings{bleurt,
	title        = {{BLEURT}: Learning Robust Metrics for Text Generation},
	author       = {Sellam, Thibault  and Das, Dipanjan  and Parikh, Ankur},
	year         = 2020,
	month        = jul,
	booktitle    = {Proceedings of the 58th Annual Meeting of the Association for Computational Linguistics},
	publisher    = {Association for Computational Linguistics},
	address      = {Online},
	pages        = {7881--7892},
	doi          = {10.18653/v1/2020.acl-main.704},
	url          = {https://aclanthology.org/2020.acl-main.704},
	editor       = {Jurafsky, Dan  and Chai, Joyce  and Schluter, Natalie  and Tetreault, Joel},
	eprint       = {2004.04696},
	archiveprefix = {arXiv},
	primaryclass = {cs.CL}
}

@inproceedings{WISEMAN2016,
	title        = {Sequence-to-Sequence Learning as Beam-Search Optimization},
	author       = {Wiseman, Sam  and Rush, Alexander M.},
	year         = 2016,
	month        = nov,
	booktitle    = {Proceedings of the 2016 Conference on Empirical Methods in Natural Language Processing},
	publisher    = {Association for Computational Linguistics},
	address      = {Austin, Texas},
	pages        = {1296--1306},
	doi          = {10.18653/v1/D16-1137},
	url          = {https://aclanthology.org/D16-1137},
	editor       = {Su, Jian  and Duh, Kevin  and Carreras, Xavier}
}

@inproceedings{ahmadian2024basics,
	title        = {Back to Basics: Revisiting {REINFORCE}-Style Optimization for Learning from Human Feedback in {LLM}s},
	author       = {Ahmadian, Arash  and Cremer, Chris  and Gall{\'e}, Matthias  and Fadaee, Marzieh  and Kreutzer, Julia  and Pietquin, Olivier  and {\"U}st{\"u}n, Ahmet  and Hooker, Sara},
	year         = 2024,
	month        = aug,
	booktitle    = {Proceedings of the 62nd Annual Meeting of the Association for Computational Linguistics (Volume 1: Long Papers)},
	publisher    = {Association for Computational Linguistics},
	address      = {Bangkok, Thailand},
	pages        = {12248--12267},
	doi          = {10.18653/v1/2024.acl-long.662},
	url          = {https://aclanthology.org/2024.acl-long.662},
	editor       = {Ku, Lun-Wei  and Martins, Andre  and Srikumar, Vivek}
}

@inproceedings{kingma2017adam,
	title        = {Adam: {A} Method for Stochastic Optimization},
	author       = {Diederik P. Kingma and Jimmy Ba},
	year         = 2015,
	booktitle    = {3rd International Conference on Learning Representations, {ICLR} 2015, San Diego, CA, USA, May 7-9, 2015, Conference Track Proceedings},
	url          = {http://arxiv.org/abs/1412.6980},
	editor       = {Yoshua Bengio and Yann LeCun},
	bibsource    = {dblp computer science bibliography, https://dblp.org}
}

@article{gemmateam2024gemma2improvingopen,
	title        = {Gemma 2: Improving Open Language Models at a Practical Size},
	author       = {Morgane Rivière and Shreya Pathak and Pier Giuseppe Sessa and Cassidy Hardin and Surya Bhupatiraju and Léonard Hussenot and Thomas Mesnard and Bobak Shahriari and Alexandre Ramé and Johan Ferret and Peter Liu and Pouya Tafti and Abe Friesen and Michelle Casbon and Sabela Ramos and Ravin Kumar and Charline Le Lan and Sammy Jerome and Anton Tsitsulin and Nino Vieillard and Piotr Stanczyk and Sertan Girgin and Nikola Momchev and Matt Hoffman and Shantanu Thakoor and Jean-Bastien Grill and Behnam Neyshabur and Olivier Bachem and Alanna Walton and Aliaksei Severyn and Alicia Parrish and Aliya Ahmad and Allen Hutchison and Alvin Abdagic and Amanda Carl and Amy Shen and Andy Brock and Andy Coenen and Anthony Laforge and Antonia Paterson and Ben Bastian and Bilal Piot and Bo Wu and Brandon Royal and Charlie Chen and Chintu Kumar and Chris Perry and Chris Welty and Christopher A. Choquette-Choo and Danila Sinopalnikov and David Weinberger and Dimple Vijaykumar and Dominika Rogozinska and Dustin Herbison and Elisa Bandy and Emma Wang and Eric Noland and Erica Moreira and Evan Senter and Evgenii Eltyshev and Francesco Visin and Gabriel Rasskin and Gary Wei and Glenn Cameron and Gus Martins and Hadi Hashemi and Hanna Klimczak-Plucinska and Harleen Batra and Harsh Dhand and Ivan Nardini and Jacinda Mein and Jack Zhou and James Svensson and Jeff Stanway and Jetha Chan and Jin Peng Zhou and Joana Carrasqueira and Joana Iljazi and Jocelyn Becker and Joe Fernandez and Joost van Amersfoort and Josh Gordon and Josh Lipschultz and Josh Newlan and Ju-yeong Ji and Kareem Mohamed and Kartikeya Badola and Kat Black and Katie Millican and Keelin McDonell and Kelvin Nguyen and Kiranbir Sodhia and Kish Greene and Lars Lowe Sjösund and Lauren Usui and Laurent Sifre and Lena Heuermann and Leticia Lago and Lilly McNealus},
	year         = 2024,
	journal      = {CoRR},
	volume       = {abs/2408.00118},
	url          = {https://doi.org/10.48550/arXiv.2408.00118},
	publtype     = {informal},
	cdate        = 1704067200000
}

@inproceedings{CPO,
	title        = {Contrastive Preference Optimization: Pushing the Boundaries of LLM Performance in Machine Translation},
	author       = {Haoran Xu and Amr Sharaf and Yunmo Chen and Weiting Tan and Lingfeng Shen and Benjamin Van Durme and Kenton Murray and Young Jin Kim},
	year         = 2024,
	booktitle    = {ICML},
	url          = {https://openreview.net/forum?id=51iwkioZpn},
	cdate        = 1704067200000
}

@inproceedings{sutskever2014sequence,
	title        = {Sequence to Sequence Learning with Neural Networks},
	author       = {Sutskever, Ilya and Vinyals, Oriol and Le, Quoc V},
	year         = 2014,
	booktitle    = {Advances in Neural Information Processing Systems},
	publisher    = {Curran Associates, Inc.},
	volume       = 27,
	pages        = {},
	url          = {https://proceedings.neurips.cc/paper_files/paper/2014/file/a14ac55a4f27472c5d894ec1c3c743d2-Paper.pdf},
	editor       = {Z. Ghahramani and M. Welling and C. Cortes and N. Lawrence and K.Q. Weinberger}
}

@inproceedings{jang2023personalizedsoupspersonalizedlarge,
	title        = {Personalized Soups: Personalized Large Language Model Alignment via Post-hoc Parameter Merging},
	author       = {Joel Jang and Seungone Kim and Bill Yuchen Lin and Yizhong Wang and Jack Hessel and Luke Zettlemoyer and Hannaneh Hajishirzi and Yejin Choi and Prithviraj Ammanabrolu},
	year         = 2024,
	booktitle    = {Adaptive Foundation Models: Evolving AI for Personalized and Efficient Learning},
	url          = {https://openreview.net/forum?id=EMrnoPRvxe}
}

@article{PPO,
	title        = {{Proximal Policy Optimization Algorithms}},
	author       = {{Schulman}, John and {Wolski}, Filip and {Dhariwal}, Prafulla and {Radford}, Alec and {Klimov}, Oleg},
	year         = 2017,
	month        = jul,
	journal      = {arXiv e-prints},
	pages        = {arXiv:1707.06347},
	doi          = {10.48550/arXiv.1707.06347},
	eid          = {arXiv:1707.06347},
	archiveprefix = {arXiv},
	eprint       = {1707.06347},
	primaryclass = {cs.LG},
	adsurl       = {https://ui.adsabs.harvard.edu/abs/2017arXiv170706347S}
}

@inproceedings{schulman2018highdimensionalcontinuouscontrolusing,
	title        = {High-Dimensional Continuous Control Using Generalized Advantage Estimation},
	author       = {John Schulman and Philipp Moritz and Sergey Levine and Michael I. Jordan and Pieter Abbeel},
	year         = 2016,
	booktitle    = {4th International Conference on Learning Representations, {ICLR} 2016, San Juan, Puerto Rico, May 2-4, 2016, Conference Track Proceedings},
	url          = {http://arxiv.org/abs/1506.02438},
	editor       = {Yoshua Bengio and Yann LeCun},
	bibsource    = {dblp computer science bibliography, https://dblp.org}
}

@inproceedings{cho2014properties,
	title        = {On the Properties of Neural Machine Translation: Encoder{--}Decoder Approaches},
	author       = {Cho, Kyunghyun  and van Merri{\"e}nboer, Bart  and Bahdanau, Dzmitry  and Bengio, Yoshua},
	year         = 2014,
	month        = oct,
	booktitle    = {Proceedings of {SSST}-8, Eighth Workshop on Syntax, Semantics and Structure in Statistical Translation},
	publisher    = {Association for Computational Linguistics},
	address      = {Doha, Qatar},
	pages        = {103--111},
	doi          = {10.3115/v1/W14-4012},
	url          = {https://aclanthology.org/W14-4012},
	editor       = {Wu, Dekai  and Carpuat, Marine  and Carreras, Xavier  and Vecchi, Eva Maria}
}

@inproceedings{ouyang2022training,
	title        = {Training language models to follow instructions with human feedback},
	author       = {Ouyang, Long and Wu, Jeffrey and Jiang, Xu and Almeida, Diogo and Wainwright, Carroll and Mishkin, Pamela and Zhang, Chong and Agarwal, Sandhini and Slama, Katarina and Ray, Alex and Schulman, John and Hilton, Jacob and Kelton, Fraser and Miller, Luke and Simens, Maddie and Askell, Amanda and Welinder, Peter and Christiano, Paul F and Leike, Jan and Lowe, Ryan},
	year         = 2022,
	booktitle    = {Advances in Neural Information Processing Systems},
	publisher    = {Curran Associates, Inc.},
	volume       = 35,
	pages        = {27730--27744},
	url          = {https://proceedings.neurips.cc/paper_files/paper/2022/file/b1efde53be364a73914f58805a001731-Paper-Conference.pdf},
	editor       = {S. Koyejo and S. Mohamed and A. Agarwal and D. Belgrave and K. Cho and A. Oh}
}

@article{flores,
	title        = {Scaling neural machine translation to 200 languages},
	author       = {{NLLB Team} and {Costa-juss{\`a}}, Marta R. and {Cross}, James and {{\c{C}}elebi}, Onur and {Elbayad}, Maha and {Heafield}, Kenneth and {Heffernan}, Kevin and {Kalbassi}, Elahe and {Lam}, Janice and {Licht}, Daniel and {Maillard}, Jean and {Sun}, Anna and {Wang}, Skyler and {Wenzek}, Guillaume and {Youngblood}, Al and {Akula}, Bapi and {Barrault}, Loic and {Mejia Gonzalez}, Gabriel and {Hansanti}, Prangthip and {Hoffman}, John and {Jarrett}, Semarley and {Sadagopan}, Kaushik Ram and {Rowe}, Dirk and {Spruit}, Shannon and {Tran}, Chau and {Andrews}, Pierre and {Fazil Ayan}, Necip and {Bhosale}, Shruti and {Edunov}, Sergey and {Fan}, Angela and {Gao}, Cynthia and {Goswami}, Vedanuj and {Guzm{\'a}n}, Francisco and {Koehn}, Philipp and {Mourachko}, Alexandre and {Ropers}, Christophe and {Saleem}, Safiyyah and {Schwenk}, Holger and {Wang}, Jeff},
	year         = 2024,
	journal      = {Nature},
	volume       = 630,
	pages        = {841--846},
	url          = {https://doi.org/10.1038/s41586-024-07335-x}
}

@article{xcomet,
	title        = {{xcomet: Transparent Machine Translation Evaluation through Fine-grained Error Detection}},
	author       = {Guerreiro, Nuno M. and Rei, Ricardo and Stigt, Daan van and Coheur, Luisa and Colombo, Pierre and Martins, André F. T.},
	year         = 2024,
	month        = {09},
	journal      = {Transactions of the Association for Computational Linguistics},
	volume       = 12,
	pages        = {979--995},
	doi          = {10.1162/tacl_a_00683},
	issn         = {2307-387X},
	url          = {https://doi.org/10.1162/tacl\_a\_00683},
	eprint       = {https://direct.mit.edu/tacl/article-pdf/doi/10.1162/tacl\_a\_00683/2468704/tacl\_a\_00683.pdf}
}

@inproceedings{DPO,
	title        = {Direct Preference Optimization: Your Language Model is Secretly a Reward Model},
	author       = {Rafailov, Rafael and Sharma, Archit and Mitchell, Eric and Manning, Christopher D and Ermon, Stefano and Finn, Chelsea},
	year         = 2023,
	booktitle    = {Advances in Neural Information Processing Systems},
	publisher    = {Curran Associates, Inc.},
	volume       = 36,
	pages        = {53728--53741},
	url          = {https://proceedings.neurips.cc/paper_files/paper/2023/file/a85b405ed65c6477a4fe8302b5e06ce7-Paper-Conference.pdf},
	editor       = {A. Oh and T. Naumann and A. Globerson and K. Saenko and M. Hardt and S. Levine}
}

@inproceedings{cometkiwi,
	title        = {Scaling up {C}omet{K}iwi: Unbabel-{IST} 2023 Submission for the Quality Estimation Shared Task},
	author       = {Rei, Ricardo  and Guerreiro, Nuno M.  and Pombal, Jos{\~A}{\copyright}  and van Stigt, Daan  and Treviso, Marcos  and Coheur, Luisa  and C. de Souza, Jos{\'e} G.  and Martins, Andr{\'e}},
	year         = 2023,
	month        = dec,
	booktitle    = {Proceedings of the Eighth Conference on Machine Translation},
	publisher    = {Association for Computational Linguistics},
	address      = {Singapore},
	pages        = {841--848},
	doi          = {10.18653/v1/2023.wmt-1.73},
	url          = {https://aclanthology.org/2023.wmt-1.73},
	editor       = {Koehn, Philipp  and Haddow, Barry  and Kocmi, Tom  and Monz, Christof}
}

@inproceedings{shen2016minimumrisktrainingneural,
	title        = {Minimum Risk Training for Neural Machine Translation},
	author       = {Shen, Shiqi  and Cheng, Yong  and He, Zhongjun  and He, Wei  and Wu, Hua  and Sun, Maosong  and Liu, Yang},
	year         = 2016,
	month        = aug,
	booktitle    = {Proceedings of the 54th Annual Meeting of the Association for Computational Linguistics (Volume 1: Long Papers)},
	publisher    = {Association for Computational Linguistics},
	address      = {Berlin, Germany},
	pages        = {1683--1692},
	doi          = {10.18653/v1/P16-1159},
	url          = {https://aclanthology.org/P16-1159},
	editor       = {Erk, Katrin  and Smith, Noah A.}
}

@inproceedings{wu2023finegrained,
	title        = {Fine-Grained Human Feedback Gives Better Rewards for Language Model Training},
	author       = {Wu, Zeqiu and Hu, Yushi and Shi, Weijia and Dziri, Nouha and Suhr, Alane and Ammanabrolu, Prithviraj and Smith, Noah A and Ostendorf, Mari and Hajishirzi, Hannaneh},
	year         = 2023,
	booktitle    = {Advances in Neural Information Processing Systems},
	publisher    = {Curran Associates, Inc.},
	volume       = 36,
	pages        = {59008--59033},
	url          = {https://proceedings.neurips.cc/paper_files/paper/2023/file/b8c90b65739ae8417e61eadb521f63d5-Paper-Conference.pdf},
	editor       = {A. Oh and T. Naumann and A. Globerson and K. Saenko and M. Hardt and S. Levine}
}

@inproceedings{RS_RATI,
	title        = {Exploration-Guided Reward Shaping for Reinforcement Learning under Sparse Rewards},
	author       = {Devidze, Rati and Kamalaruban, Parameswaran and Singla, Adish},
	year         = 2022,
	booktitle    = {Advances in Neural Information Processing Systems},
	publisher    = {Curran Associates, Inc.},
	volume       = 35,
	pages        = {5829--5842},
	url          = {https://proceedings.neurips.cc/paper_files/paper/2022/file/266c0f191b04cbbbe529016d0edc847e-Paper-Conference.pdf},
	editor       = {S. Koyejo and S. Mohamed and A. Agarwal and D. Belgrave and K. Cho and A. Oh}
}

@inproceedings{RS_GOYAL,
	title        = {Using Natural Language for Reward Shaping in Reinforcement Learning},
	author       = {Goyal, Prasoon and Niekum, Scott and Mooney, Raymond J.},
	year         = 2019,
	month        = 7,
	booktitle    = {Proceedings of the Twenty-Eighth International Joint Conference on Artificial Intelligence, {IJCAI-19}},
	publisher    = {International Joint Conferences on Artificial Intelligence Organization},
	pages        = {2385--2391},
	doi          = {10.24963/ijcai.2019/331},
	url          = {https://doi.org/10.24963/ijcai.2019/331}
}

@inproceedings{Tower,
	title        = {Tower: An Open Multilingual Large Language Model for Translation-Related Tasks},
	author       = {Duarte Miguel Alves and Jos{\'e} Pombal and Nuno M Guerreiro and Pedro Henrique Martins and Jo{\~a}o Alves and Amin Farajian and Ben Peters and Ricardo Rei and Patrick Fernandes and Sweta Agrawal and Pierre Colombo and Jos{\'e} G. C. de Souza and Andre Martins},
	year         = 2024,
	booktitle    = {First Conference on Language Modeling},
	url          = {https://openreview.net/forum?id=EHPns3hVkj}
}

@inproceedings{WMT24,
	title        = {Findings of the {WMT}24 General Machine Translation Shared Task: The {LLM} Era Is Here but {MT} Is Not Solved Yet},
	author       = {Kocmi, Tom  and Avramidis, Eleftherios  and Bawden, Rachel  and Bojar, Ond{\v{r}}ej  and Dvorkovich, Anton  and Federmann, Christian  and Fishel, Mark  and Freitag, Markus  and Gowda, Thamme  and Grundkiewicz, Roman  and Haddow, Barry  and Karpinska, Marzena  and Koehn, Philipp  and Marie, Benjamin  and Monz, Christof  and Murray, Kenton  and Nagata, Masaaki  and Popel, Martin  and Popovi{\'c}, Maja  and Shmatova, Mariya  and Steingr{\'\i}msson, Steinth{\'o}r  and Zouhar, Vil{\'e}m},
	year         = 2024,
	month        = nov,
	booktitle    = {Proceedings of the Ninth Conference on Machine Translation},
	publisher    = {Association for Computational Linguistics},
	address      = {Miami, Florida, USA},
	pages        = {1--46},
	doi          = {10.18653/v1/2024.wmt-1.1},
	url          = {https://aclanthology.org/2024.wmt-1.1},
	editor       = {Haddow, Barry  and Kocmi, Tom  and Koehn, Philipp  and Monz, Christof}
}

@inproceedings{colombo2022bestsystemsnewperspectives,
	title        = {What are the best Systems? New Perspectives on NLP Benchmarking},
	author       = {Colombo, Pierre and Noiry, Nathan and Irurozki, Ekhine and Cl\'{e}men\c{c}on, Stephan},
	year         = 2022,
	booktitle    = {Advances in Neural Information Processing Systems},
	publisher    = {Curran Associates, Inc.},
	volume       = 35,
	pages        = {26915--26932},
	url          = {https://proceedings.neurips.cc/paper_files/paper/2022/file/ac4920f4085b5662133dd751493946a6-Paper-Conference.pdf},
	editor       = {S. Koyejo and S. Mohamed and A. Agarwal and D. Belgrave and K. Cho and A. Oh}
}

@article{BPE,
	title        = {A new algorithm for data compression},
	author       = {Philip Gage},
	year         = 1994,
	journal      = {The C Users Journal archive},
	volume       = 12,
	pages        = {23--38},
	url          = {https://api.semanticscholar.org/CorpusID:59804030}
}

@inproceedings{fa,
	title        = {FlashAttention: Fast and Memory-Efficient Exact Attention with {IO}-Awareness},
	author       = {Tri Dao and Daniel Y Fu and Stefano Ermon and Atri Rudra and Christopher Re},
	year         = 2022,
	booktitle    = {Advances in Neural Information Processing Systems},
	url          = {https://openreview.net/forum?id=H4DqfPSibmx},
	editor       = {Alice H. Oh and Alekh Agarwal and Danielle Belgrave and Kyunghyun Cho}
}

@inproceedings{fa2,
	title        = {FlashAttention-2: Faster Attention with Better Parallelism and Work Partitioning},
	author       = {Tri Dao},
	year         = 2024,
	booktitle    = {The Twelfth International Conference on Learning Representations},
	url          = {https://openreview.net/forum?id=mZn2Xyh9Ec}
}

@inproceedings{quant,
	title        = {Quantization and Training of Neural Networks for Efficient Integer-Arithmetic-Only Inference},
	author       = {Jacob, Benoit and Kligys, Skirmantas and Chen, Bo and Zhu, Menglong and Tang, Matthew and Howard, Andrew and Adam, Hartwig and Kalenichenko, Dmitry},
	year         = 2018,
	month        = {June},
	booktitle    = {Proceedings of the IEEE Conference on Computer Vision and Pattern Recognition (CVPR)}
}

@article{dist,
	title        = {{Distilling the Knowledge in a Neural Network}},
	author       = {{Hinton}, Geoffrey and {Vinyals}, Oriol and {Dean}, Jeff},
	year         = 2015,
	month        = mar,
	journal      = {arXiv e-prints},
	pages        = {arXiv:1503.02531},
	doi          = {10.48550/arXiv.1503.02531},
	eid          = {arXiv:1503.02531},
	archiveprefix = {arXiv},
	eprint       = {1503.02531},
	primaryclass = {stat.ML},
	adsurl       = {https://ui.adsabs.harvard.edu/abs/2015arXiv150302531H}
}

@inproceedings{WuSeq2017,
	title        = {Sequence Prediction with Unlabeled Data by Reward Function Learning},
	author       = {Lijun Wu and Li Zhao and Tao Qin and Jianhuang Lai and Tie-Yan Liu},
	year         = 2017,
	booktitle    = {Proceedings of the Twenty-Sixth International Joint Conference on Artificial Intelligence, {IJCAI-17}},
	pages        = {3098--3104},
	doi          = {10.24963/ijcai.2017/432},
	url          = {https://doi.org/10.24963/ijcai.2017/432}
}

@inproceedings{Bahdanau2016,
	title        = {An Actor-Critic Algorithm for Sequence Prediction},
	author       = {Dzmitry Bahdanau and Philemon Brakel and Kelvin Xu and Anirudh Goyal and Ryan Lowe and Joelle Pineau and Aaron Courville and Yoshua Bengio},
	year         = 2017,
	booktitle    = {International Conference on Learning Representations},
	url          = {https://openreview.net/forum?id=SJDaqqveg}
}

@inproceedings{li-etal-2024-reinforcement,
	title        = {Reinforcement Learning with Token-level Feedback for Controllable Text Generation},
	author       = {Li, Wendi  and Wei, Wei  and Xu, Kaihe  and Xie, Wenfeng  and Chen, Dangyang  and Cheng, Yu},
	year         = 2024,
	month        = jun,
	booktitle    = {Findings of the Association for Computational Linguistics: NAACL 2024},
	publisher    = {Association for Computational Linguistics},
	address      = {Mexico City, Mexico},
	pages        = {1704--1719},
	doi          = {10.18653/v1/2024.findings-naacl.111},
	url          = {https://aclanthology.org/2024.findings-naacl.111/},
	editor       = {Duh, Kevin  and Gomez, Helena  and Bethard, Steven}
}

@inproceedings{xia-etal-2024-inverse,
	title        = {Inverse-{Q}*: Token Level Reinforcement Learning for Aligning Large Language Models Without Preference Data},
	author       = {Xia, Han  and Gao, Songyang  and Ge, Qiming  and Xi, Zhiheng  and Zhang, Qi  and Huang, Xuanjing},
	year         = 2024,
	month        = nov,
	booktitle    = {Findings of the Association for Computational Linguistics: EMNLP 2024},
	publisher    = {Association for Computational Linguistics},
	address      = {Miami, Florida, USA},
	pages        = {8178--8188},
	doi          = {10.18653/v1/2024.findings-emnlp.478},
	url          = {https://aclanthology.org/2024.findings-emnlp.478/},
	editor       = {Al-Onaizan, Yaser  and Bansal, Mohit  and Chen, Yun-Nung}
}

@inproceedings{yoon-etal-2024-tlcr,
	title        = {{TLCR}: Token-Level Continuous Reward for Fine-grained Reinforcement Learning from Human Feedback},
	author       = {Yoon, Eunseop  and Yoon, Hee Suk  and Eom, SooHwan  and Han, Gunsoo  and Nam, Daniel  and Jo, Daejin  and On, Kyoung-Woon  and Hasegawa-Johnson, Mark  and Kim, Sungwoong  and Yoo, Chang},
	year         = 2024,
	month        = aug,
	booktitle    = {Findings of the Association for Computational Linguistics: ACL 2024},
	publisher    = {Association for Computational Linguistics},
	address      = {Bangkok, Thailand},
	pages        = {14969--14981},
	doi          = {10.18653/v1/2024.findings-acl.889},
	url          = {https://aclanthology.org/2024.findings-acl.889/},
	editor       = {Ku, Lun-Wei  and Martins, Andre  and Srikumar, Vivek}
}

@inproceedings{2023arXiv231104072G,
	title        = {Beyond Imitation: Leveraging Fine-grained Quality Signals for Alignment},
	author       = {Geyang Guo and Ranchi Zhao and Tianyi Tang and Xin Zhao and Ji{-}Rong Wen},
	year         = 2024,
	booktitle    = {The Twelfth International Conference on Learning Representations, {ICLR} 2024, Vienna, Austria, May 7-11, 2024},
	url          = {https://openreview.net/forum?id=LNLjU5C5dK},
	bibsource    = {dblp computer science bibliography, https://dblp.org}
}

@article{2024arXiv241100722O,
	title        = {{Token-level Proximal Policy Optimization for Query Generation}},
	author       = {{Ouyang}, Yichen and {Wang}, Lu and {Yang}, Fangkai and {Zhao}, Pu and {Huang}, Chenghua and {Liu}, Jianfeng and {Pang}, Bochen and {Yang}, Yaming and {Zhan}, Yuefeng and {Sun}, Hao and {Lin}, Qingwei and {Rajmohan}, Saravan and {Deng}, Weiwei and {Zhang}, Dongmei and {Sun}, Feng and {Zhang}, Qi},
	year         = 2024,
	month        = nov,
	journal      = {arXiv e-prints},
	pages        = {arXiv:2411.00722},
	doi          = {10.48550/arXiv.2411.00722},
	eid          = {arXiv:2411.00722},
	archiveprefix = {arXiv},
	eprint       = {2411.00722},
	primaryclass = {cs.LG},
	adsurl       = {https://ui.adsabs.harvard.edu/abs/2024arXiv241100722O}
}

@inproceedings{cao-etal-2024-enhancing,
	title        = {Enhancing Reinforcement Learning with Dense Rewards from Language Model Critic},
	author       = {Cao, Meng  and Shu, Lei  and Yu, Lei  and Zhu, Yun  and Wichers, Nevan  and Liu, Yinxiao  and Meng, Lei},
	year         = 2024,
	month        = nov,
	booktitle    = {Proceedings of the 2024 Conference on Empirical Methods in Natural Language Processing},
	publisher    = {Association for Computational Linguistics},
	address      = {Miami, Florida, USA},
	pages        = {9119--9138},
	doi          = {10.18653/v1/2024.emnlp-main.515},
	url          = {https://aclanthology.org/2024.emnlp-main.515/},
	editor       = {Al-Onaizan, Yaser  and Bansal, Mohit  and Chen, Yun-Nung}
}

@inproceedings{lu-etal-2024-error,
	title        = {Error Analysis Prompting Enables Human-Like Translation Evaluation in Large Language Models},
	author       = {Lu, Qingyu  and Qiu, Baopu  and Ding, Liang  and Zhang, Kanjian  and Kocmi, Tom  and Tao, Dacheng},
	year         = 2024,
	month        = aug,
	booktitle    = {Findings of the Association for Computational Linguistics: ACL 2024},
	publisher    = {Association for Computational Linguistics},
	address      = {Bangkok, Thailand},
	pages        = {8801--8816},
	doi          = {10.18653/v1/2024.findings-acl.520},
	url          = {https://aclanthology.org/2024.findings-acl.520/},
	editor       = {Ku, Lun-Wei  and Martins, Andre  and Srikumar, Vivek}
}

@inproceedings{ramos-aligning,
	title        = {Aligning Neural Machine Translation Models: Human Feedback in Training and Inference},
	author       = {Ramos, Miguel Moura  and Fernandes, Patrick  and Farinhas, Ant{\'o}nio  and Martins, Andre},
	year         = 2024,
	month        = jun,
	booktitle    = {Proceedings of the 25th Annual Conference of the European Association for Machine Translation (Volume 1)},
	publisher    = {European Association for Machine Translation (EAMT)},
	address      = {Sheffield, UK},
	pages        = {258--274},
	url          = {https://aclanthology.org/2024.eamt-1.22/},
	editor       = {Scarton, Carolina  and Prescott, Charlotte  and Bayliss, Chris  and Oakley, Chris  and Wright, Joanna  and Wrigley, Stuart  and Song, Xingyi  and Gow-Smith, Edward  and Bawden, Rachel  and S{\'a}nchez-Cartagena, V{\'i}ctor M  and Cadwell, Patrick  and Lapshinova-Koltunski, Ekaterina  and Cabarr{\~a}o, Vera  and Chatzitheodorou, Konstantinos  and Nurminen, Mary  and Kanojia, Diptesh  and Moniz, Helena}
}
